\documentclass{article}

\usepackage[sglblindworkshop, final]{neurips_2025}

\usepackage[utf8]{inputenc} 
\usepackage[T1]{fontenc}    
\usepackage{hyperref}       
\usepackage{url}            
\usepackage{booktabs}       
\usepackage{amsfonts}       
\usepackage{nicefrac}       
\usepackage{microtype}      
\usepackage{xcolor}         
\usepackage{amsmath} 
\usepackage{graphicx}
\usepackage{subcaption}
\usepackage{float}

\workshoptitle{AI4Mat-NeurIPS 2025}

\title{Superior Molecular Representations from Intermediate Encoder Layers}

\author{%
  Luis Pinto\\
  \texttt{pinto.luisc@gmail.com} \\
}

\begin{document}

\maketitle

\begin{abstract}


Pretrained molecular encoders have become indispensable in computational chemistry for tasks such as property prediction and molecular generation. However, the standard practice of relying solely on final-layer embeddings for downstream tasks may discard valuable information. In this work, we first analyze the information flow in five diverse molecular encoders and find that intermediate layers retain more general-purpose features, whereas the final-layer specializes and compresses information. We then perform an empirical layer-wise evaluation across 22 property prediction tasks. We find that using frozen embeddings from optimal intermediate layers improves downstream performance by an average of 5.4\%, up to 28.6\%, compared to the final-layer. Furthermore, finetuning encoders truncated at intermediate depths achieves even greater average improvements of 8.5\%, with increases as high as 40.8\%, obtaining new state-of-the-art results on several benchmarks. These findings highlight the importance of exploring the full representational depth of molecular encoders to achieve substantial performance improvements and computational efficiency. The code is made publicly available at \href{https://github.com/luispintoc/Unlocking-Chemical-Insights}{https://github.com/luispintoc/Unlocking-Chemical-Insights}.

\end{abstract}

\section{Introduction}

Deep learning has reshaped molecular science, where pretrained molecular encoders are essential tools for applications from virtual screening in drug discovery to designing novel materials with desired properties \cite{goh2018smiles2vecinterpretablegeneralpurposedeep, sypetkowski2024on, doi:10.1021/acs.chemrestox.5c00018}. These encoders, built on architectures such as Transformers \cite{vaswani2023attentionneed} and Graph Neural Networks (GNNs) \cite{NIPS2017_5dd9db5e, kipf2017semisupervisedclassificationgraphconvolutional}, learn rich representations of molecular structures that can be applied to a wide range of downstream predictive tasks. The common practice of extracting molecular representations from the final encoder layer, while simple and widely adopted, rests on the implicit assumption that this layer consistently provides the most informative and task-relevant features.

However, this assumption is increasingly challenged by findings from Natural Language Processing (NLP) and Computer Vision, where intermediate layers often encode richer, more generalizable, or task-specific information, leading to significant performance gains \cite{liu-etal-2019-linguistic, jin-etal-2025-exploring, gurnee2024languagemodelsrepresentspace,valeriani2023geometryhiddenrepresentationslarge}. Despite the unique data modalities (e.g. molecular graphs, 3D conformers) and distinct pretraining objectives inherent to cheminformatics, systematic layer-wise studies of molecular encoders have been notably scarce. This leaves a critical question unanswered: are we overlooking superior molecular representations by adhering to the final-layer convention in chemistry?

\begin{figure}[t]
  \centering
  \begin{subfigure}[t]{0.48\textwidth}
    \centering
    \includegraphics[width=\linewidth]{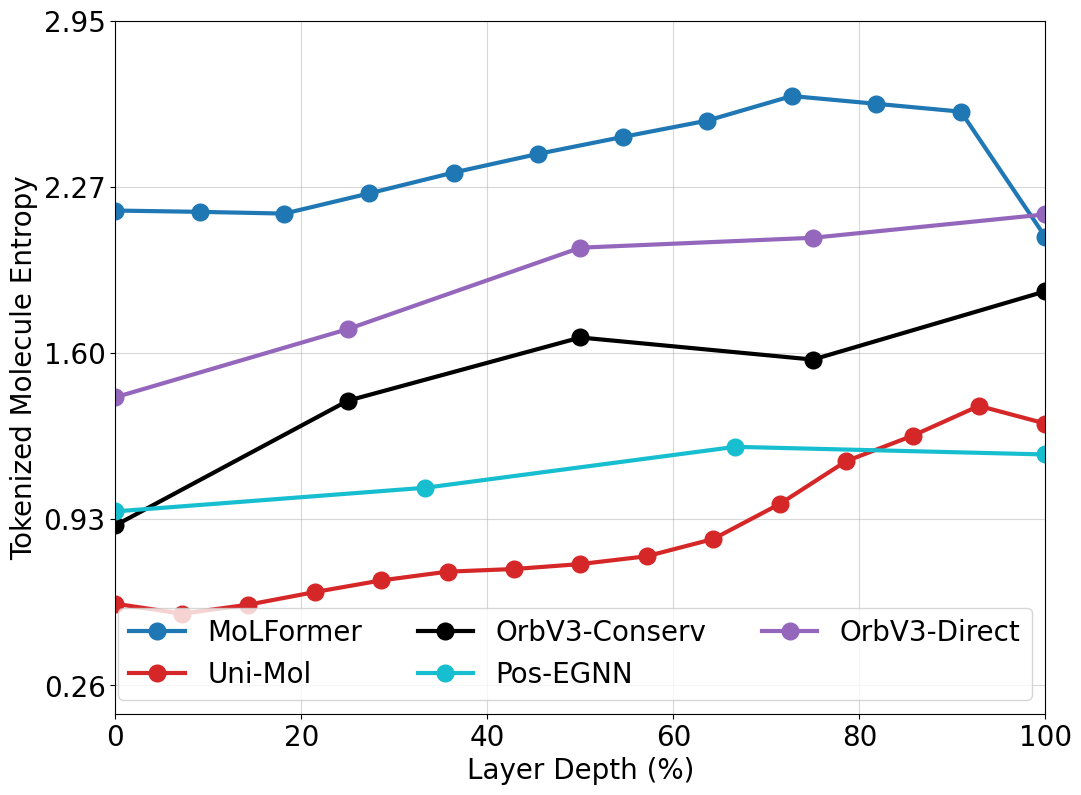}
    \subcaption{Tokenized-molecule Entropy}
    \label{fig:tme}
  \end{subfigure}\hfill
  \begin{subfigure}[t]{0.48\textwidth}
    \centering
    \includegraphics[width=\linewidth]{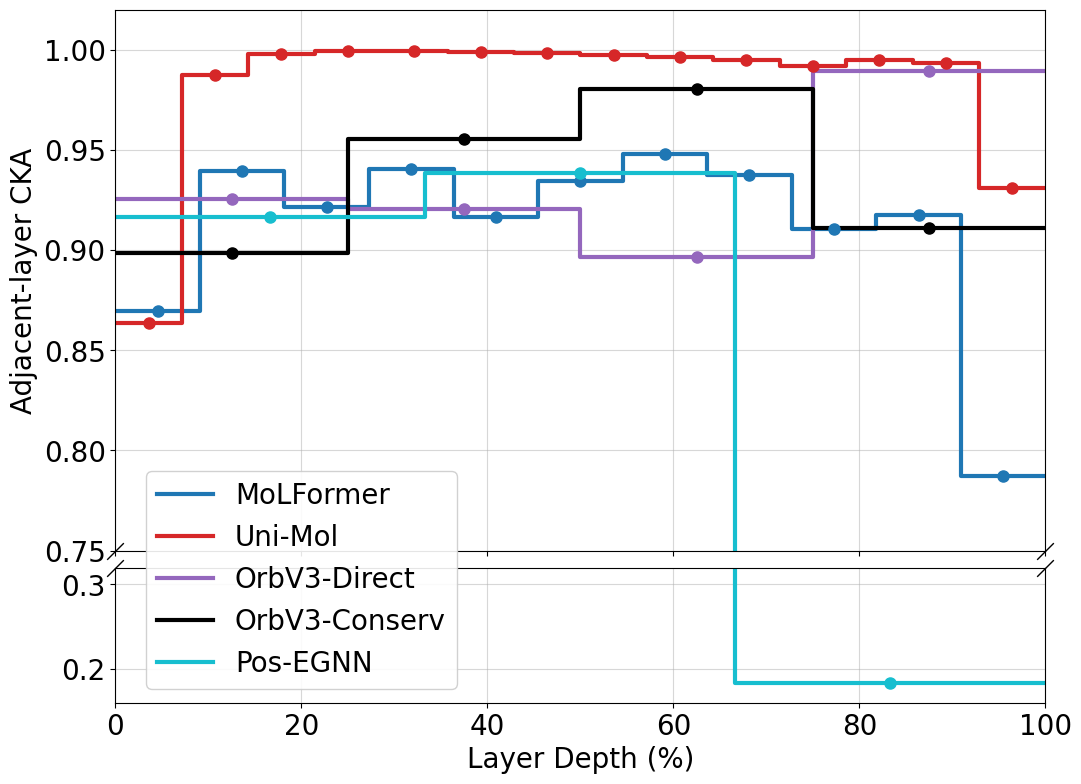}
    \subcaption{Adjacent-layer CKA}
    \label{fig:adjacent-cka}
  \end{subfigure}
  \caption{Left: Tokenized-molecule entropy rises with depth but falls at the final block for MolFormer, Uni-Mol and PosEGNN, indicating compression; Orb models maintain higher spread. Right: Adjacent-layer CKA shows small changes across interior layers and a pronounced last-step change for most models. Depth is normalized from first encoder block (0\%) to last (100\%).}
  \label{fig:information}

\end{figure}

Our work consists of a two-stage analysis across five diverse pretrained encoders. First, we  map the information flow through their hidden layers, and second, we empirically evaluate the performance of these layer-wise representations on a suite of downstream tasks.


\newpage

Our main contributions are as follows:
\begin{enumerate}
    \item We use two label-free probes, tokenized-molecule entropy and adjacent-layer linear centered kernel alignment (CKA), to characterize information flow within molecules and across model depth. Our analysis reveals that while the middle layers remain stable and capture general, transferable features, the pronounced change in the final layer signals both information compression and specialization for the pretraining task.
    \item We conduct an empirical study evaluating frozen embeddings across 22 absorption, distribution, metabolism, excretion, and toxicity (ADMET) tasks. We demonstrate that in over 81\% of model-task combinations, an intermediate layer outperforms the conventionally used final-layer, yielding an average increase of 5.4\% in downstream performance.
    \item We further show that these advantages persist when the encoders are partially finetuned. For the same set of tasks, we observe that finetuning up to an optimal intermediate layer leads to notable improvements over finetuning the complete encoder in over 71\% of cases, with an average gain of 8.5\%.
\end{enumerate}

\section{Related Work}

\subsection{Molecular Encoders}

Molecular encoders are deep learning models designed to transform chemical structures into fixed-length numerical vectors, also known as embeddings, that can be used for downstream tasks such as property prediction, molecular generation, or virtual screening. These models play a central role in computational chemistry and drug discovery by enabling machine learning algorithms to efficiently capture the relevant chemical, structural, and physical information inherent to molecules. The effectiveness of a molecular encoder is critically dependent on both the representational richness of its learned embeddings and its ability to generalize across diverse chemical spaces.

Molecular encoders are generally differentiated by two main aspects: the type of molecular input they process, for example SMILES \cite{doi:10.1021/ci00057a005} strings, molecular graphs, or 3D conformers; and the underlying computational architecture, which could be Transformers, Graph Neural Networks (GNNs), or other specialized architectures.

\textbf{Transformer-based sequence models} such as ChemBERTa \cite{chithrananda2020chembertalargescaleselfsupervisedpretraining}, ChemGPT \cite{frey_soklaski_axelrod_samsi_gomez-bombarelli_coley_gadepally_2022}, and MolFormer \cite{Ross2022} are pretrained on textual representations of molecules and apply masked language modeling \cite{devlin-etal-2019-bert} or autoregressive objectives \cite{Radford2019LanguageMA} to learn general purpose embeddings. ChemBERTa and MolFormer operate on tokenized SMILES strings, whereas ChemGPT is trained on SELFIES \cite{Krenn_2020}, a robust string grammar that guarantees valid molecules. These models inherit the architectural benefits of NLP transformers, including multi-head self-attention and position embeddings.

\textbf{Hybrid 2D/3D transformer encoders} like Uni-Mol \cite{zhou2023unimol} extend transformer architectures to directly incorporate both 2D topological information and 3D atomic coordinates. Uni-Mol leverages pairwise distance matrices and spatial positional encodings to inject geometric priors, enabling the model to learn both conformation-aware and topology-sensitive representations. Its architecture retains the global receptive field of transformers while being sensitive to the spatial layout of molecules.

\textbf{Graph Neural Networks for 2D molecular graphs} represent a distinct branch of molecular representation learning. These models operate directly on the 2D graph structure of molecules, where atoms are nodes and bonds are edges. Through iterative message passing steps, GNNs aggregate information from neighboring atoms and bonds to learn representations that capture molecular topology and chemical features. Models like the Directed Message Passing Neural Network (D-MPNN) \cite{HAN2022100201} are often trained with supervised learning on specific molecular properties. Others, such as MolCLR \cite{Wang2022}, employ self-supervised learning strategies, like contrastive learning between positive and negative molecular pairs, to learn general purpose embeddings.

\textbf{GNN-based 3D models} employ graph neural networks to process three dimensional molecular structures. Some of these models incorporate equivariant architectures to inherently respect physical symmetries, while others may utilize data augmentation techniques to learn these invariances. A specific subset of these 3D GNN models includes machine learning force fields \cite{Behler2016-og, doi:10.1021/acs.chemrev.0c01111}, which are trained to predict molecular energies and forces from DFT calculations \cite{PhysRev.140.A1133}. For instance, MACE \cite{Batatia2022mace} and Pos-EGNN \cite{Pos-EGNN} are examples that use an equivariant architecture to predict such properties. In contrast, Orb models \cite{rhodes2025orbv3atomisticsimulationscale} do not rely on equivariant architectures but instead leverage augmentations.

\subsection{Layer-Wise Probing in Language and Vision Models}

Most relevant to our study is the recent work of Skean et\,al.\,(2025) \cite{skean2025layerlayeruncoveringhidden}, who performed the most comprehensive analysis to date of intermediate-layer representations in deep neural networks. Their layer-by-layer investigation across transformer and state-space models in both language and vision domains revealed that intermediate layers often exhibit higher utility than final ones across a range of probing tasks. Using a unified framework integrating metrics from information theory, geometry, and invariance, they demonstrated that the most informative representations, outperforming final-layer embeddings by up to 16\% on 32 tasks, tend to emerge midway through the model. At this stage, task-relevant signals are preserved while over-specialization and noise accumulation, potentially more prevalent in final-layers, have not yet occurred. Their work critically challenges the widely adopted practice of extracting representations solely from the final-layer and uncovers how different architectural and training paradigms shape internal information flow and compression.

Despite these advances, layer-wise representational analysis has not been applied to molecular encoders, which differ from NLP and vision models in both data modality and pretraining objectives. Molecular encoders often combine structural and chemical priors, and their performance is often assessed using challenging benchmarks that include both regression and classification tasks, operate with limited data, or employ scaffold splits to test distinct generalization capabilities. Yet nearly all such models, from SMILES transformers to equivariant GNNs, default to using the final-layer as the molecular embedding, without examining the structure or quality of intermediate representations.

\subsection{Surrogate Evaluators for Embedding Quality}

Evaluating the quality of learned representations without resorting to computationally expensive end-to-end finetuning is crucial for efficient model development and analysis. This has led to the use of various surrogate models to probe the information encoded in frozen embeddings. These surrogates provide an estimate of the downstream utility across different tasks.

Linear probes, such as logistic regression or linear regression, are frequently used to assess the linear separability of concepts within embedding spaces \cite{Tripepi2008}. More complex, tree-based ensemble models such as Random Forests or Gradient Boosted Decision Trees (e.g., XGBoost \cite{chen_xgboost_2016}, LightGBM \cite{ke_lightgbm_2017}, CatBoost \cite{dorogush2018catboostgradientboostingcategorical}) are often effective when applied to frozen embeddings \cite{priyadarsini2024improvingperformancepredictionelectrolyte, soares2025representing}. In some cases, small Multi-Layer Perceptrons (MLPs) are also trained as probes \cite{Li2023, Masood2025}. More recently, models leveraging in-context learning principles have demonstrated significant promise. TabPFN \cite{hollmann2023tabpfntransformersolvessmall}, built upon a transformer architecture and pretrained on vast amounts of synthetic tabular data, can make accurate predictions without requiring task-specific gradient updates or hyperparameter tuning. These surrogate evaluators enable efficient assessment of learned embeddings by pairing powerful pretrained encoders with lightweight predictors. From linear models to in-context transformer models, they offer a scalable alternative to finetuning, balancing speed and predictive reliability for downstream evaluation.

\section{Methods}

This section details the molecular encoder architectures investigated, the benchmark datasets and evaluation metrics employed, the information flow analysis and the experimental protocols for evaluating layer-wise representations through both frozen embeddings and finetuning.

\subsection{Molecular Encoder Models}
\label{section 3.1}

Our study investigates the layer-wise representations from five diverse pretrained molecular encoder architectures, selected to cover various input modalities and model designs relevant to molecular representation learning. These include:

\begin{itemize}
    \item MolFormer \cite{Ross2022}: A transformer architecture pretrained on tokenized SMILES strings with a masked language modeling objective. This model consists of 12 layers.
    \item Uni-Mol \cite{zhou2023unimol}: We evaluated the first model in the Uni-Mol family, which extends transformer architectures to directly incorporate both 2D topological information and 3D atomic coordinates. It features 15 encoder layers.

    \item Orb Family \cite{rhodes2025orbv3atomisticsimulationscale}: We investigated two specific variants from the Orb family of machine learning force fields. They do not rely explicitly on equivariant message passing architectures but incorporate invariances through their data augmentation. The variants we used are:
    \begin{itemize}
        \item Orb-v3-conservative-omat: This variant is trained with a "conservative" objective, where forces are derived as the negative gradient of the predicted energy. The checkpoint used was trained on the ab initio molecular dynamics subset of OMat24 \cite{barrosoluque2024openmaterials2024omat24}. It features 5 interaction layers.
        \item Orb-v3-direct-mpa: This variant is trained with a "direct" objective to predict forces explicitly, alongside energy. This checkpoint was trained on the combination of MPTraj \cite{10.1063/1.4812323} and Alexandria (PBE) \cite{Schmidt2023-af} datasets. It also features 5 interaction layers.
    \end{itemize}
    \item Pos-EGNN \cite{Pos-EGNN}: A Position-based Equivariant Graph Neural Network (Pos-EGNN) that functions as a machine learning force field. This foundation model for chemistry and materials utilizes equivariant GNNs operating directly on 3D molecular conformers. Representations were extracted from its 4 equivariant message passing blocks, which we treat as distinct layers in our analysis.
    
\end{itemize}

To generate 3D conformations for all molecules, we employed the ETKDG algorithm \cite{doi:10.1021/acs.jcim.5b00654}. These conformations were subsequently optimized using the MMFF94 force field \cite{HalgrenThomasA.1996Mmff} to ensure physically plausible geometries before input into models requiring 3D coordinates. It should be noted that fewer than five data points per task (<\,0.5\% of the total datasets) did not yield valid 3D conformers using this protocol, and these instances were excluded from downstream analysis.


Finally, we emphasize that while the Orb models and Pos-EGNN are machine learning force fields trained on inorganic datasets, we investigated their learned representations here in an organic property prediction context for the first time. As these models were not trained for standard QSAR or ADMET prediction, our evaluation provides insight into the generality and transferability of their intermediate representations beyond their original design objectives.

\subsection{Benchmark and Evaluation Protocol}
\label{section 3.2}

We evaluated molecular encoder representations on the TDCommons (TDC) benchmark \cite{Huang2021tdc}, a curated suite of 22 ADMET-related tasks spanning both regression and classification settings. Each task requires the prediction of a single target, a specific physicochemical or biological property, using molecular structures as input. ADMET properties critically influence the development and safety profiles of pharmaceutical compounds; thus, maximizing representational quality has tangible benefits for reducing experimental costs and enhancing predictive accuracy in drug discovery \cite{doi:10.1021/acs.chemrestox.5c00273, MacDermottOpeskin2025ChemRxiv}.

To address the challenge of generalizing to novel chemical scaffolds, TDC provides scaffold-based data splits that partition each dataset into training, validation, and testing sets. This ensures that compounds in the test set have distinct core structures from those in the training set, enabling a rigorous assessment of a model’s capacity to extrapolate to unseen chemical space.

We aggregated the training partitions across tasks to construct an unlabeled corpus for the information flow analysis. For the empirical experiments, we used the prescribed splits per task without modification. Following TDC benchmark guidelines, performance for each task was quantified using its designated metric. For regression tasks, this metric was either Mean Absolute Error (MAE) or Spearman Rank Correlation Coefficient, while for classification tasks, it was either the Area Under the Receiver Operating Characteristic curve (AUROC) or the Area Under the Precision-Recall Curve (AUCPR). Thus, our comprehensive evaluation framework employs these four performance metrics. Further details on each task are provided in Appendix\,\ref{appendix A}.

\subsection{Information Flow Probes}
\label{section 3.3}

We use two label-free probes to diagnose how information evolves across layers and to test the hypothesis that the final-layer is often over-specialized for pretraining. Let \(x_i\) denote a molecule and \(\ell\) an encoder layer. The token matrix is \(H_{i,\ell}\in\mathbb{R}^{T_i\times d}\), where \(T_i\) is the number of valid tokens/atoms for molecule \(x_i\) and \(d\) is the hidden dimensionality of layer \(\ell\). To derive a single vector representation per molecule from each layer, we followed the conventions established in the original publications of the respective models. For instance, MolFormer uses mean pooling over token embeddings, while Uni-Mol utilizes the representation of the special [CLS] token. For GNN-based models such as Pos-EGNN and the Orb variants, mean pooling over all scalar node embeddings was employed to obtain the graph-level representation.

\paragraph{Tokenized-molecule entropy.} We first quantify the diversity of within-molecule token signals with a matrix-based entropy \cite{giraldo2014measuresentropydatausing}. For each \(x_i\) and layer \(\ell\), we form the token Gram \(K_{i,\ell}=H_{i,\ell}H_{i,\ell}^\top\), take its eigenvalues \(\{\lambda_t\}\), normalize \(p_t=\lambda_t/\sum_u \lambda_u\), and average the Shannon entropy of this spectrum across molecules:
\[
\mathrm{TME}(\ell)=\frac{1}{|\mathcal D|}\sum_{i\in\mathcal D}\Big(-\sum_t p_t\log p_t\Big).
\]
Higher values indicate that token embeddings are spread across many principal directions, expressing greater variety and lower redundancy. Lower values indicate that variance concentrates in a small number of directions, reflecting higher redundancy and representational compression, a collapse towards a low-rank subspace.

\paragraph{Adjacent-layer CKA on pooled molecule vectors.}
We assess changes in the pooled space by comparing adjacent layers using linear centered kernel alignment (CKA) \cite{kornblith2019similarityneuralnetworkrepresentations}. For two adjacent layers \(\ell\) and \(\ell{+}1\), we form \(X\) by stacking, row-wise, the pooled molecule vectors from layer \(\ell\) (one row per molecule), and likewise \(Y\) from layer \(\ell{+}1\). We then center each matrix across examples by subtracting the corresponding row-mean vector, yielding \(\tilde X\) and \(\tilde Y\). The centered second moments are then defined as
\[
S_{xx}^{(\ell)}=\tilde X^\top \tilde X,\qquad
S_{yy}^{(\ell+1)}=\tilde Y^\top \tilde Y,\qquad
S_{xy}^{(\ell,\ell+1)}=\tilde X^\top \tilde Y.
\]
And the linear CKA between layers is
\[
\mathrm{CKA}_{\ell\to \ell+1}
=\frac{\|S_{xy}^{(\ell,\ell+1)}\|_F^2}{\|S_{xx}^{(\ell)}\|_F\,\|S_{yy}^{(\ell+1)}\|_F}\in[0,1],
\]

Values of \(\mathrm{CKA}_{\ell\to \ell+1}\) close to 1 indicate that consecutive layers produce highly similar molecule embeddings, while values closer to 0 indicate larger geometric changes between layers. We report adjacent-layer CKA across depth to quantify, in a label-free and scale-free manner, how representation changes attenuate or persist throughout the encoder.

\subsection{Evaluating Frozen Layer-wise Embeddings} \label{section 3.4}

To assess the predictive utility of representations from different layers, we use the hidden states from every encoder layer as frozen molecule embeddings for each TDC task. As explained in Section \ref{section 3.3}, we obtained a single vector representation for each molecule at every layer by following the conventions outlined in the original publications of the corresponding models.

These fixed-length embeddings then served as input to TabPFNv2 \cite{hollmann2025tabpfn}, hereafter simply TabPFN, a transformer-based in-context learning model pretrained on a vast array of synthetic tabular data. We selected TabPFN as our lightweight downstream predictive model for the following key reasons:
\begin{enumerate}
    \item \textbf{No Hyperparameter Tuning:} TabPFN makes predictions for new tasks in a single forward pass without requiring task-specific hyperparameter optimization. This was crucial for our study, as it allows for a fair comparison of embedding quality across different layers and models, minimizing the confound of surrogate model tuning.
    \item \textbf{Strong Validated Performance:} TabPFN has demonstrated state-of-the-art performance on diverse small to medium-scale tabular classification and regression tasks, making it a robust choice for efficiently probing embedding utility. Furthermore, as detailed in Appendix\,\ref{appendix B}, our own preliminary experiments confirmed that its performance when using cheminformatic features as input is comparable to that of well-established tree-based models on the same tasks, further validating its suitability as a reliable and efficient evaluator when applied to learned embeddings in this analysis.
\end{enumerate}


For every TDC task, we retrieved the embeddings produced by each individual layer, trained TabPFN on the scaffold‑split training portion of those embeddings and their labels, generated predictions for the matching test embeddings, and scored them with the metrics from Section\,\ref{section 3.2}.

\subsection{Finetuning with Layer-wise Representations}

We also performed a comprehensive finetuning study across all pretrained molecular encoders and TDC tasks. For each encoder and task, we extracted molecular representations from individual encoder layers, using the same pooling strategy as Section\,\ref{section 3.4}, and fed them into a task-specific prediction head. This prediction head consisted of two linear layers: the first projected to a hidden dimension, followed by a SiLU activation  \cite{ELFWING20183}, 10\% dropout, and a final linear layer projecting to the output dimension.

For each specific encoder layer being evaluated on each task, key hyperparameters, namely the prediction head's hidden dimension selected from \{32, 256\} and the learning rate for the AdamW optimizer \cite{loshchilov2019decoupledweightdecayregularization} selected from \{1e-5, 2e-5, 5e-5, 1e-4, 2e-4\}, were optimized via an independent hyperparameter search. Hyperparameter ranges were selected based on preliminary studies and align with configurations commonly used in the original publications of the respective models, effectively capturing performance variance without excessive computational demands.

During training, both the parameters of the encoder up to and including the selected layer and those of the prediction head were jointly finetuned; any encoder layers beyond the selected layer were excluded. Models were trained for 50 epochs with a fixed batch size of 64 and the learning rate was linearly warmed up over the first 5\% steps and decayed over the remaining steps. Validation performance was monitored at each epoch, and the model checkpoint yielding the best validation metric was used for test set evaluation.


\section{Results and Discussion}

In this section, we begin with a label-free analysis of representation dynamics using the probes in Section~\ref{section 3.3}. These depth-wise diagnostics reveal how token diversity and layer-to-layer geometry evolve inside each encoder. We then validate these findings by evaluating model performance across 22 TDC ADMET tasks in two settings. We first show that intermediate layer representations consistently outperform the final-layer when used as inputs to TabPFN. We then show that these advantages largely persist when the encoders are finetuned up to an optimal intermediate layer. Finally, we establish a correlation between these two evaluation paradigms, suggesting an efficient strategy for selecting optimal layers for finetuning.

\subsection{Unsupervised Probes of Representation Dynamics}

Figure \ref{fig:information} summarizes depth-wise behavior across models using two label-free probes. First, an analysis of token diversity using tokenized-molecule entropy reveals a pattern of end-of-encoder compression in several models. MolFormer and Uni-Mol exhibit a sharp drop in entropy in their final-layer, with Pos-EGNN showing a milder version of this effect. This drop indicates that the token representations collapse into a low-rank subspace, concentrating their variance into fewer principal directions. Such behavior is consistent with the final-layer becoming highly tailored to its pretraining objective. Conversely, both Orb variants maintain a broader spread across principal directions, indicating that token information stays more distributed instead of collapsing at the end.

Second, geometric stability between adjacent layers, captured by CKA, reveals that Uni-Mol sits near 1 across depth, indicating very similar pooled molecule embeddings from one block to the next. The remaining encoders live in the 0.90 to 0.95 range, which points to small but meaningful updates as depth increases. Most models then show a pronounced last-step drop in CKA signaling a large geometric remapping right before the output, that aligns with the pretraining task specialization and coincides with the token-level compression above. Viewed jointly, these trends support the view that, regardless of the specific architecture or pretraining objective, the final-layer is often over-specialized and does not always carry the most transferable features.

\begin{figure}[!t]
    \centering
    \makebox[\textwidth][c]{\includegraphics[width=1.0\linewidth]{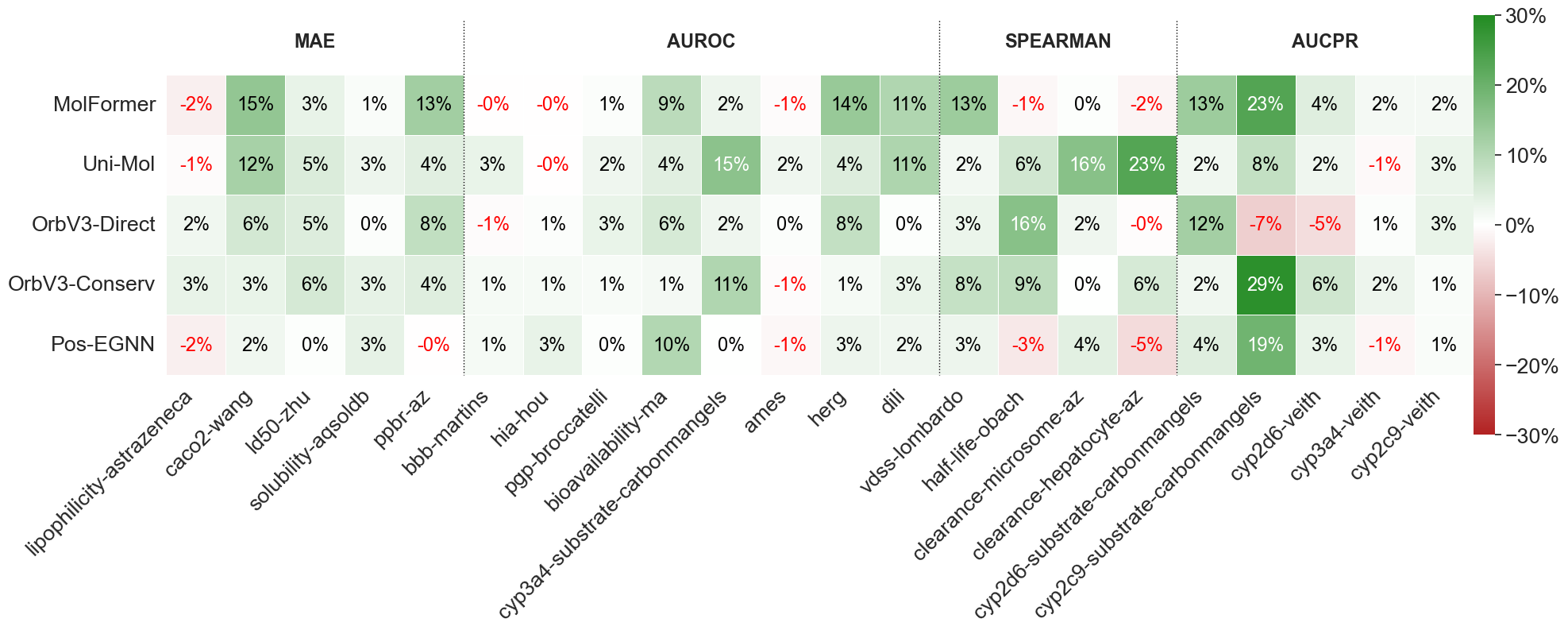}}
    \caption{Percentage improvement in test metric of the best intermediate layer relative to the final-layer when evaluating frozen embeddings. Positive values mean the best non-final layer improves over the final-layer. Negative values mean the final-layer outperforms the best non-final layer.}
    \label{fig:heatmap1}
    \vspace{-3mm}
\end{figure}

\subsection{Superior Performance of Intermediate Layer Embeddings}

We evaluate downstream utility by using each layer’s hidden states as fixed feature embeddings with TabPFN. Figure~\ref{fig:heatmap1} reports, for each model and task pair, the signed percent change of the best non-final layer relative to the final-layer, where green values indicate gains over the final-layer and red values indicate that the final-layer is better.

Across all evaluations, 81\% of model and task combinations prefer a non-final layer. Negative values are present, although they are small in magnitude; when the final-layer wins, its average margin over the best non-final layer is about 1\%. The choice of encoder influences these gains. MolFormer and Uni-Mol benefit the most from intermediate layers, with average improvements of 7.9\% and 6.7\%, respectively. In contrast, the comparatively shallow models show an average improvement of only $\sim$4\%.  

Moreover, while prior work in language and vision reports that the most informative representations often emerge midway through the model \cite{skean2025layerlayeruncoveringhidden}, our empirical study did not identify a universally optimal depth for molecular encoders. Instead, the ideal layer varies by architecture and ADMET task, as illustrated by the full performance curves in Appendix~\ref{appendix C}. This pattern aligns with the information flow findings. Across the interior of each encoder, adjacent-layer CKA is high and nearly constant, indicating that successive layers produce very similar pooled molecule vectors; in practice this means many intermediate layers supply comparably good features, so no single depth consistently dominates. In contrast, between the final two layers we observe a pronounced CKA drop, and tokenized-molecule entropy shows a concurrent decrease, both consistent with a last-stage geometric remapping and increased compression tailored to the pretraining objective. In aggregate, these observations explain why intermediate layers frequently outperform the final-layer while the precise winning depth varies by model–task pair.

Task-specific benefits were also evident. For instance, \textit{cyp2c9-substrate-carbonmangels} saw substantial improvements from intermediate layers, with average gains of 19.8\%. Conversely, tasks like \textit{clearance-hepatocyte-az} and \textit{ames} showed limited advantage, with intermediate layers outperforming the final-layer in only two of the five models examined. Furthermore, using intermediate layers also yielded new state-of-the-art (SOTA) results, for example: Pos-EGNN Layer 0 on \textit{hia-hou} (AUROC 0.994 vs.\ 0.989), Uni-Mol Layer 3 on \textit{clearance-microsome-az} (Spearman 0.641 vs.\ 0.630), and Pos-EGNN Layer 3 on \textit{dili} (AUROC 0.942 vs.\ 0.925) \cite{notwell2023admetpropertypredictioncombinations,Turon2022.12.13.520154}.

\subsection{Superiority of Intermediate Layers Persists with Finetuning}

\begin{figure}[!t]
    \centering
    \makebox[\textwidth][c]{\includegraphics[width=1.0\linewidth]{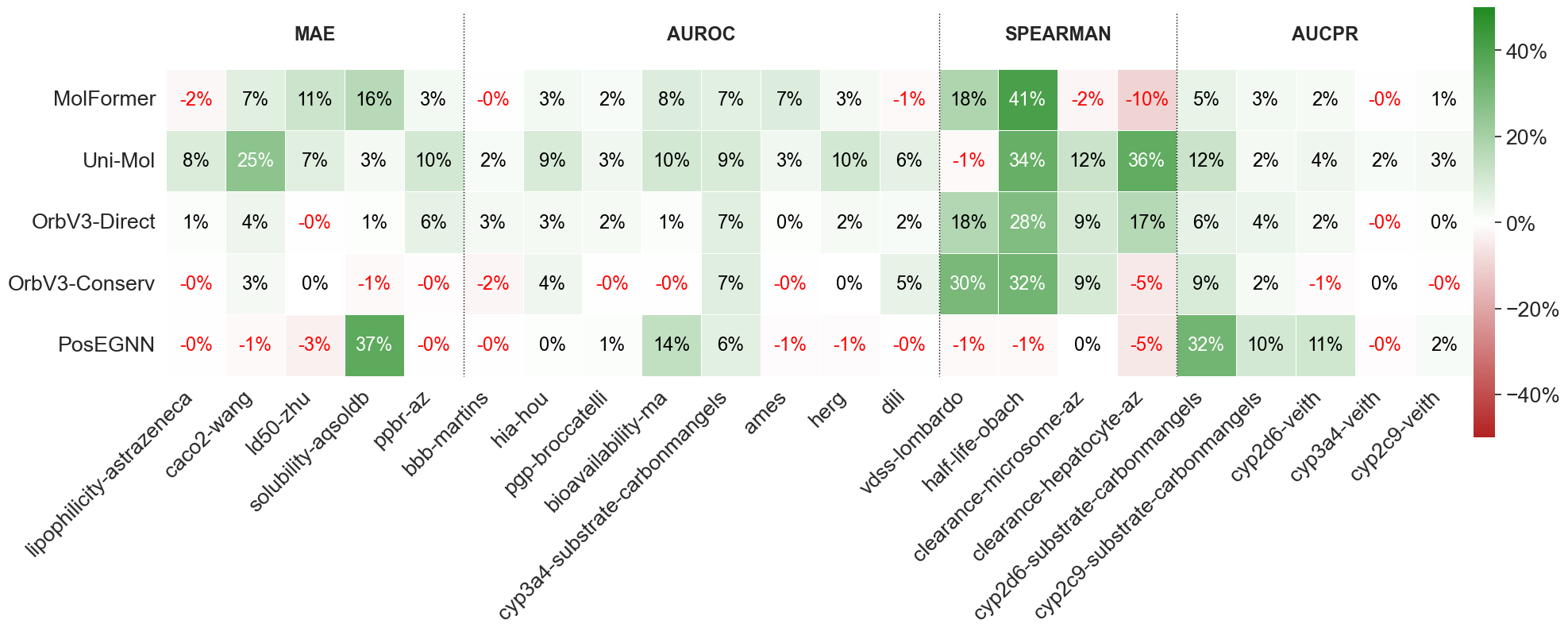}}
    \caption{Percentage change in test metric achieved by finetuning up to the best intermediate layer compared to finetuning up to the final-layer. Positive values mean the best non-final layer improves over the final-layer. Negative values mean the final-layer outperforms the best non-final layer.}
    \label{fig:heatmap2}
    \vspace{-3mm}
\end{figure}

Next, we investigated whether the observed benefits of intermediate layers translate to a scenario where the encoder, up to a selected layer, and a task-specific prediction head are jointly finetuned. Figure\,\ref{fig:heatmap2} shows the relative difference in the test metric when finetuning to the optimal intermediate layer compared to finetuning to the final-layer, again with each cell representing a specific model-task pair and showing the numerical percentage gain. Layer‐by‐layer performance visualizations for all models and tasks are provided in Appendix\,\ref{appendix C}.

We observed pronounced benefits from employing intermediate layers even in this finetuning context.  Averaged across all pairs, the optimal intermediate layer improves the test metric by 8.5\%, with 71\% of pairs showing gains. The gains were model-dependent but consistently substantial: Pos-EGNN saw the highest average improvement at 11.2\%, closely followed by Uni-Mol at 9.9\%, while MolFormer and OrbV3-Direct each experienced average gains of 8.4\%. OrbV3-Conservative exhibited smaller, yet still notable, gains of 5.8\%.
Consistent with the information flow diagnostics and frozen embedding results, no single depth was best, which point to broadly similar interior representations and a specialized final block.

Tasks evaluated with Spearman correlation benefited the most from intermediate layers (e.g., \textit{half-life-obach} +33.8\%, \textit{vdss-lombardo} +22\%), although negatives do occur and \textit{clearance-hepatocyte-az} shows both large gains and large drops across models, underscoring task-specific sensitivity to depth. Moreover, several intermediate layers deliver better results than prior SOTA benchmarks even without extensive hyperparameter tuning. For instance, Pos-EGNN Layer 3 on \textit{lipophilicity-astrazeneca} (MAE 0.459 vs.\ 0.467), MolFormer Layer 8 on \textit{ppbr-az} (MAE 7.221 vs.\ 7.526) and Uni-Mol Layer 13 on \textit{cyp3a4-substrate-carbonmangels} (AUROC 0.688 vs.\ 0.662)\cite{notwell2023admetpropertypredictioncombinations, Turon2022.12.13.520154, doi:10.1021/acs.jcim.9b00237,10.1093/bioinformatics/btaa1005}.




Beyond predictive improvements, leveraging intermediate layers substantially reduces the number of trainable parameters. Since the number of parameters scales nearly linearly with encoder depth, stopping finetuning at an intermediate layer proportionally reduces the computational burden. For instance, MolFormer truncated at layer 8 of its 12 total layers prunes the final third of the network, decreasing trainable parameters by approximately 33\%. This yields models that not only train faster but also require fewer computational resources, facilitating efficient experimentation and deployment without compromising, and frequently even enhancing, predictive performance.



\subsection{Frozen Embedding Performance as a Proxy for Finetuning Layer Selection}

\begin{figure}[!tbp]
  \centering
  %
  \begin{subfigure}[b]{0.49\textwidth}
    \centering
    \includegraphics[width=\textwidth]{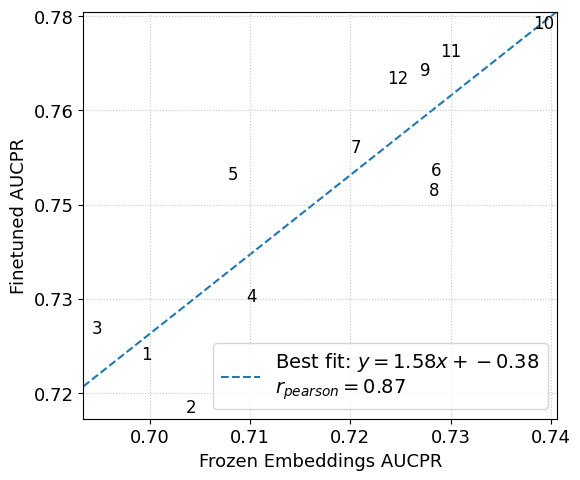}
  \end{subfigure}
  \hfill
  %
  \begin{subfigure}[b]{0.49\textwidth}
    \centering
    \includegraphics[width=\textwidth]{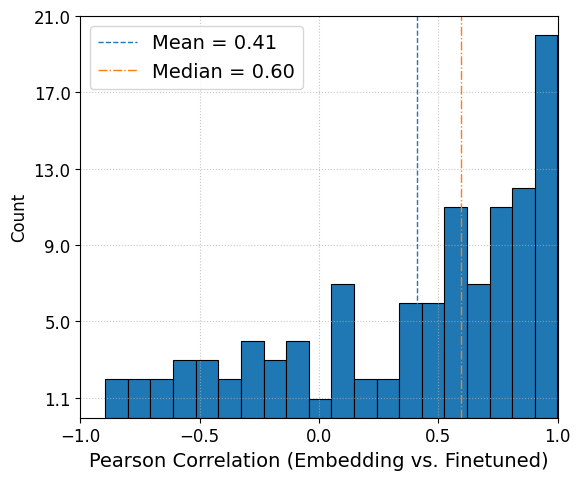}
  \end{subfigure}
  \caption{Left: Example of scatter plot of frozen embedding AUCPR vs. finetuned AUCPR for each MolFormer layer on task \textit{cyp2c9-veith}. Each point is annotated with its corresponding layer number. Right: Histogram of embedding-to-finetuned correlations for all 110 model-task combinations.}
  \label{fig:histogram}
  \vspace{-3mm}
\end{figure}

For each encoder-task pair in the TDC, we computed the Pearson correlation between the layer‑wise frozen embedding scores and the corresponding finetuned scores, thereby quantifying how well embedding performance anticipates finetuning gains. Figure\,\ref{fig:histogram} shows the correlation distribution. The left panel shows one representative scatter plot for MolFormer on the \textit{cyp2c9-veith} task, where each data point corresponds to a layer, with a Pearson correlation of 0.87. This near linear relationship indicates that layers scoring highest in frozen embedding metric are typically the same layers that achieve the top metric after finetuning. Individual scatter plots for all model-task combinations are provided in Appendix\,\ref{appendix D}\,--\,\ref{appendix H}. The right panel aggregates this analysis across all 110 model–task pairs. The distribution is heavily skewed toward positive correlations: the median is 0.60, signaling a strong alignment for half of the model‑task pairs and indicating frozen embedding scores can serve as a practical first-order filter to prioritize layers for finetuning, substantially reducing the search space.



\section{Conclusion}

This work systematically challenges the common practice of relying on final-layer representations from pretrained molecular encoders. Our comprehensive study, spanning five diverse molecular encoders and 22 ADMET tasks, demonstrates that intermediate layers frequently offer superior downstream performance. This superiority is evident when evaluating frozen embeddings, where in over 81\% of model-task combinations an intermediate layer outperformed the final-layer, yielding an average 5.4\% performance increase. With finetuning, intermediate layers led to notable improvements in over 71\% of cases, with an average gain of 8.5\%, alongside significant savings in parameters and compute time. These empirical gains are supported by our label-free analysis, which shows that the final-layer's specialization often leads to representational compression, leaving intermediate layers with more general transferable features. 




While our findings are encouraging, we acknowledge certain limitations. Although constrained by computational resources to five encoders and 22 ADMET tasks, extending this analysis to a broader set of molecular encoders and chemical domains could further validate our findings. Future work with additional architectures, multiple random seeds, and more exhaustive hyperparameter searches would provide deeper insights and more robust generalizations.  Nevertheless, our results strongly advocate for a shift beyond the final-layer default, urging exploration of the rich representational landscape within molecular encoders.


\bibliographystyle{unsrt}
\bibliography{neurips_2025}


\newpage

\section*{Appendix}  
\addcontentsline{toc}{section}{Appendix}  

\appendix  

\section{TDC-ADMET Task Summary} \label{appendix A}

Table \ref{tab:datasets} summarizes the 22 tasks from the Therapeutics Data Commons ADMET benchmark suite used in our experiments. Each row corresponds to a distinct molecular property prediction task, with associated details including a brief description, the number of compounds available in the dataset, and the evaluation metric used.

\begin{table}[!htpb]
  \centering
  \begin{tabular}{@{}p{2.5cm}p{6.6cm}rrl@{}}
    \toprule
    \textbf{Dataset}     & \textbf{Description}                                                      & \textbf{\# compounds} & \textbf{Metric}   \\
    \midrule
    lipophilicity-astrazeneca         & Measures the ability of a drug to dissolve in a lipid                                            & 4\,200  & MAE   \\
    caco2-wang          & Estimates intestinal permeability                               & 906 & MAE   \\
    ld50-zhu         & Indicates drug's lethal dose threshold                  & 7\,385  & MAE   \\
    solubility-aqsoldb      & Measures a drug's ability to dissolve in water                                                   & 9\,982  & MAE   \\
    ppbr-az      & lasma protein binding percentage influences drug delivery efficiency                                                   & 1\,614  & MAE   \\
    \midrule
    bbb-martins        & Drug penetration across the blood-brain barrier                                       & 1\,975  & AUROC   \\
    hia-hou        & Human intestinal absorption impacting oral drug delivery                                      & 578  & AUROC   \\
    pgp-broccatelli          & P-glycoprotein inhibition affects drug bioavailability, resistance                                                 & 1\,212 & AUROC  \\
    bioavailability-ma          & Oral bioavailability determines systemic drug exposure                                     & 640 & AUROC  \\
    cyp3a4-substrate-carbonmangels         & CYP3A4 metabolizes drugs for bodily clearance                                                            & 667  & AUROC      \\
    ames     & Mutagenicity assesses genetic damage via Ames test                                   & 7\,255     & AUROC      \\
    herg & hERG inhibition linked to cardiac safety risks                                & 648  & AUROC      \\
    dili & Drug-induced liver injury impacting drug safety                                & 475  & AUROC      \\
    \midrule
    vdss-lombardo & Volume of distribution reflects tissue drug concentration                                & 1\,130  & SPEARMAN      \\
    half-life-obach & Half-life indicates duration of drug activity                                & 667  & SPEARMAN      \\
    clearance-microsome-az & Drug clearance measures rate of systemic elimination                                & 1\,102  & SPEARMAN      \\
    clearance-hepatocyte-az & Drug clearance measures rate of systemic elimination                                & 1\,1020  & SPEARMAN      \\
    \midrule
    cyp2d6-substrate-carbonmangels & CYP2D6 enzyme involved in drug metabolism                                & 664  & AUCPR      \\
    cyp2c9-substrate-carbonmangels & Octanol/water distribution coefficient of molecules                                & 666  & AUCPR      \\
    cyp2d6-veith & CYP2C9 catalyzes metabolism of various compounds                                & 13\,130 & AUCPR      \\
    cyp3a4-veith & CYP3A4 oxidizes xenobiotics for drug clearance                                & 12\,328  & AUCPR      \\
    cyp2c9-veith & CYP2C9 mediates oxidation of cellular compounds                                & 12\,092  & AUCPR      \\
    \bottomrule
  \end{tabular}
  \vspace{3mm}
  \caption{Evaluated datasets description.}
  \label{tab:datasets}
\end{table}

\newpage

\section{Validating TabPFN Against a Strong Tree‑Based Baseline} \label{appendix B}

To ensure that TabPFN is an adequate surrogate for assessing embedding quality, we compared it to a state-of-the-art gradient‑boosted decision‑tree model. Specifically, we followed the CatBoost pipeline described in \cite{notwell2023admetpropertypredictioncombinations}: each molecule is represented by the concatenation of ECFP‑1024 \cite{doi:10.1021/ci100050t}, Avalon‑1024 \cite{doi:10.1021/ci050413p}, ErG‑315 \cite{doi:10.1021/ci050457y}, 200 descriptors from RDKit \cite{rdkit_zenodo}, and a 300‑dimensional GIN embedding \cite{xu2019powerfulgraphneuralnetworks}, yielding a 2863‑feature input that achieved top‑3 accuracy on 19/22 TDC‑ADMET leaderboards.

\begin{figure}[!htpb]
    \centering
    \makebox[\textwidth][c]{\includegraphics[width=0.95\linewidth]{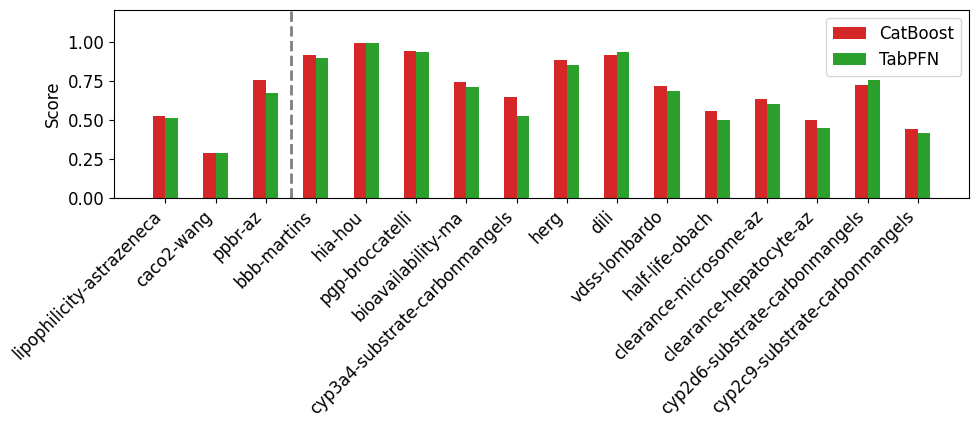}}
    \caption{Comparison of CatBoost and TabPFN across a suite of ADMET benchmarks. The vertical dashed line separates the MAE‐based tasks on the left (lower is better) from the remaining tasks on the right (higher is better). To display \textit{ppbr‐az} on a 0–1 scale, its MAE was divided by 10.}
    \label{fig:tab_vs_cat}
\end{figure}

Due to compute constraints, we evaluated TabPFN and CatBoost on only 16 of the 22 ADMET benchmarks, as shown in Figure\,\ref{fig:tab_vs_cat}. On these 16 tasks, TabPFN matched CatBoost’s overall performance and even outperformed it on five tasks. Remarkably, TabPFN achieved this with a single forward pass, no hyperparameter tuning and far more features than it was pretrained on (a maximum of 500 features). This confirms our choice of TabPFN as a lightweight, hyperparameter-free probe: it isolates the impact of encoder representations without introducing variability from per-task model optimization.

\section{Layer-wise Performance} \label{appendix C}

Each figure in this appendix presents the complete layer‑wise trajectories for all five encoders on a given TDC task. The left panel reports the frozen embedding performance obtained by freezing the encoder at successive depths and training a lightweight TabPFN surrogate on top; the right panel shows the outcome when the encoder is finetuned up to the same depth together with a task‑specific prediction head. Finetuning hyperparameters for each model and layer were determined by conducting a limited grid search over a few learning rates and hidden layer sizes. Depth is expressed as a percentage of the total number of encoder blocks (0\% = first encoder layer; 100 \% = final block), and performance is plotted according to its task metric.


As expected, the two Orb variants, pretrained on inorganic data, generally perform worse than the other models across both evaluation paradigms, highlighting the significance of domain-aligned pretraining for small-molecule property prediction. However, Pos-EGNN, trained on a subset of the OrbV3-Direct dataset, consistently demonstrates strong performance, frequently matching or surpassing models like MolFormer and Uni-Mol. This surprising outcome indicates that the lower performance of the Orb variants may not result solely from domain mismatch, but also from insufficient hyperparameter tuning.

Comparing the two panels reveals that frozen embedding curves frequently, but not always, anticipate the relative ordering observed after fine‑tuning. In some tasks (e.g. \textit{lipophilicity‑astrazeneca} and \textit{ppbr-az}) the layer that minimizes/maximizes the given metric in the surrogate setting also yields the best or near best score after finetuning; in others, the correspondence weakens, suggesting that perhaps a broader hyperparameter search is needed. Additionally, it should be noted that for a small percentage of cases, such as \textit{vdss-lombardo} and \textit{half-life-obach}, finetuning provided lower performance than using frozen embeddings. This variability likely contributes to the long tail observed in the histogram of Pearson correlations (Figure \ref{fig:histogram}).

Finally, no single region of the network consistently outperforms the rest. Optimal depths vary not only across encoders but also across tasks within the same encoder. This heterogeneity reinforces the utility of our two‑step protocol: inexpensive frozen embedding evaluation first narrows the set of candidate layers, after which targeted finetuning can focus computational resources on the most promising depths.



\begin{figure}[!htpb]
    \centering
    \makebox[\textwidth][c]{\includegraphics[width=1.0\linewidth]{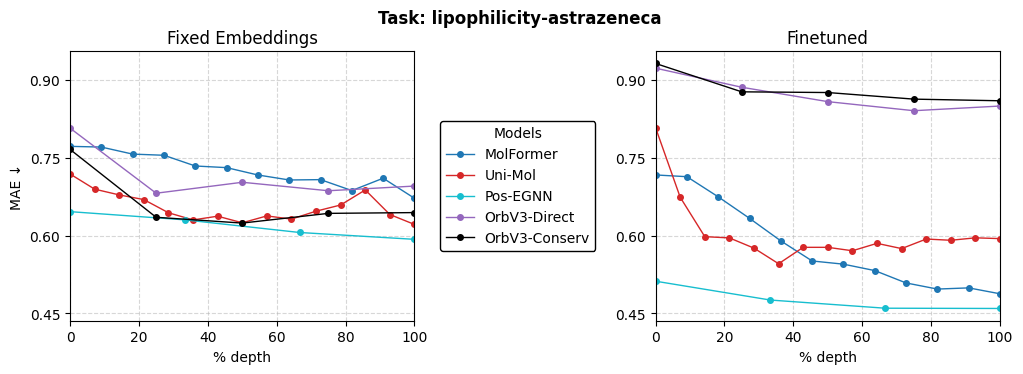}}
    \caption{MAE (↓, the lower the better) on the \textbf{\textit{lipophilicity-astrazeneca}} dataset as a function of encoder depth (\% depth) for frozen embeddings (left) versus finetuned embeddings (right).}
\end{figure}

\begin{figure}[!htpb]
    \centering
    \makebox[\textwidth][c]{\includegraphics[width=1.0\linewidth]{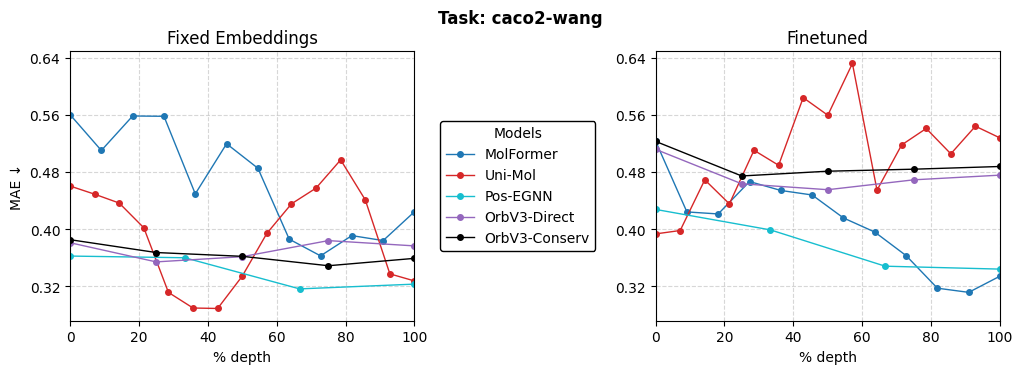}}
    \caption{MAE (↓, the lower the better) on the \textbf{\textit{caco2-wang}} dataset as a function of encoder depth (\% depth) for frozen embeddings (left) versus finetuned embeddings (right).}
\end{figure}

\begin{figure}[!htpb]
    \centering
    \makebox[\textwidth][c]{\includegraphics[width=1.0\linewidth]{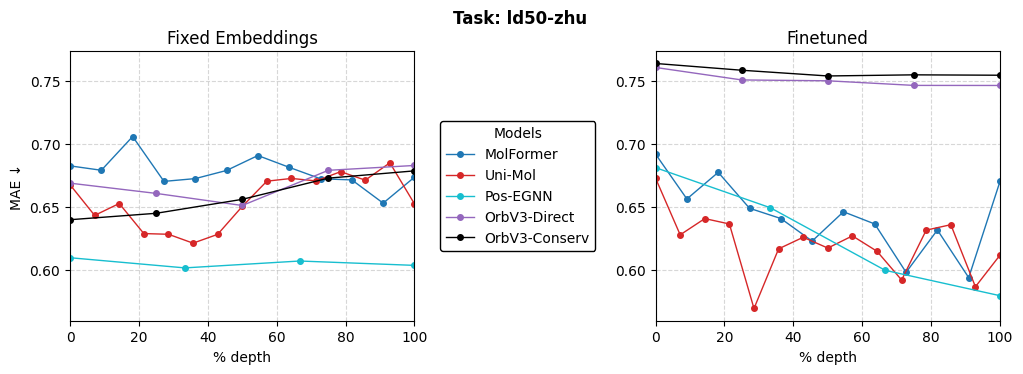}}
    \caption{MAE (↓, the lower the better) on the \textbf{\textit{ld50-zhu}} dataset as a function of encoder depth (\% depth) for frozen embeddings (left) versus finetuned embeddings (right).}
\end{figure}

\begin{figure}[!htpb]
    \centering
    \makebox[\textwidth][c]{\includegraphics[width=1.0\linewidth]{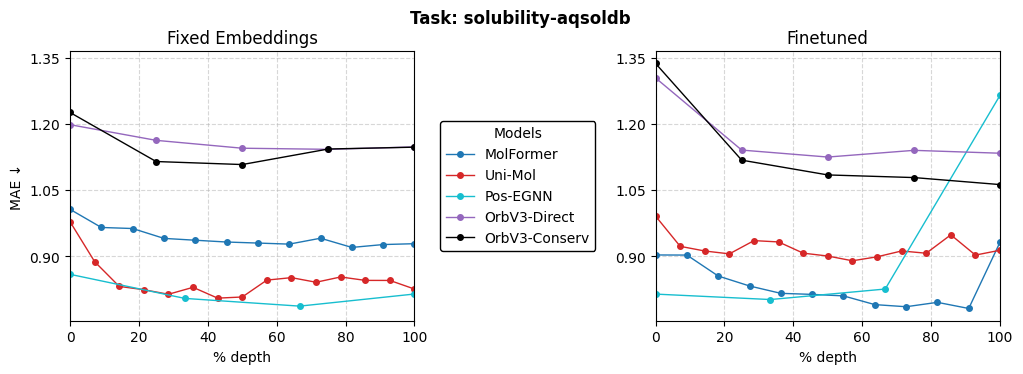}}
    \caption{MAE (↓, the lower the better) on the \textbf{\textit{solubility-aqsoldb}} dataset as a function of encoder depth (\% depth) for frozen embeddings (left) versus finetuned embeddings (right).}
\end{figure}

\begin{figure}[!htpb]
    \centering
    \makebox[\textwidth][c]{\includegraphics[width=1.0\linewidth]{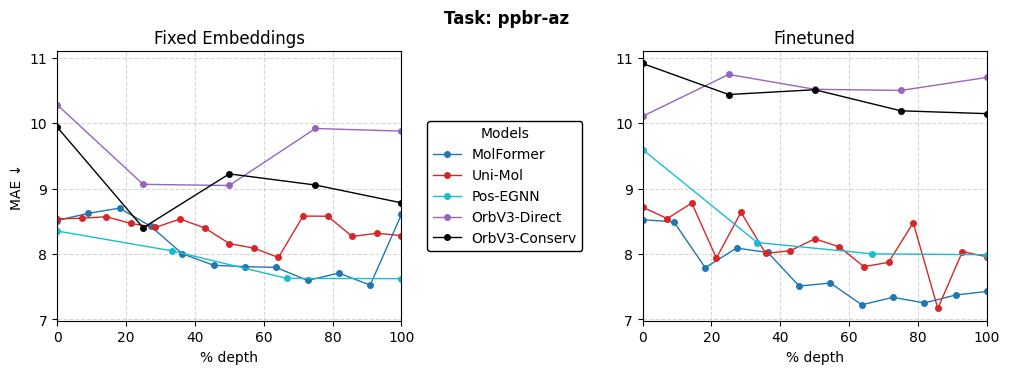}}
    \caption{MAE (↓, the lower the better) on the \textbf{\textit{ppbr-az}} dataset as a function of encoder depth (\% depth) for frozen embeddings (left) versus finetuned embeddings (right).}
\end{figure}

\begin{figure}[!htpb]
    \centering
    \makebox[\textwidth][c]{\includegraphics[width=1.0\linewidth]{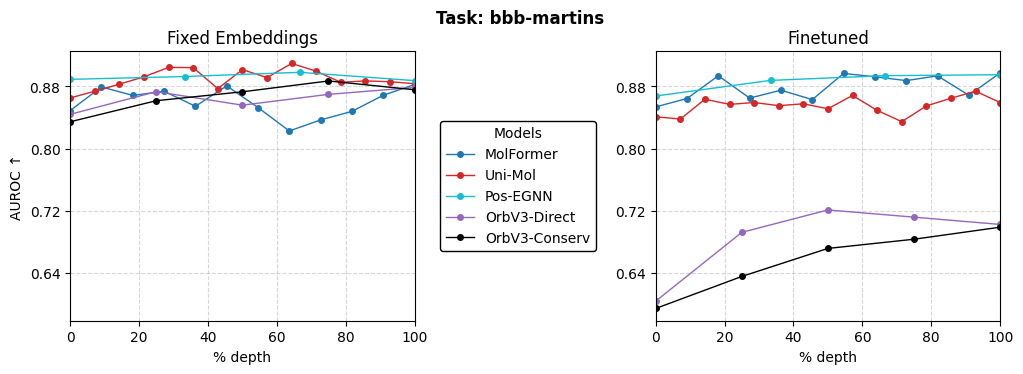}}
    \caption{AUROC (↑, the higher the better) on the \textbf{\textit{bbb-martins}} dataset as a function of encoder depth (\% depth) for frozen embeddings (left) versus finetuned embeddings (right).}
\end{figure}

\begin{figure}[!htpb]
    \centering
    \makebox[\textwidth][c]{\includegraphics[width=1.0\linewidth]{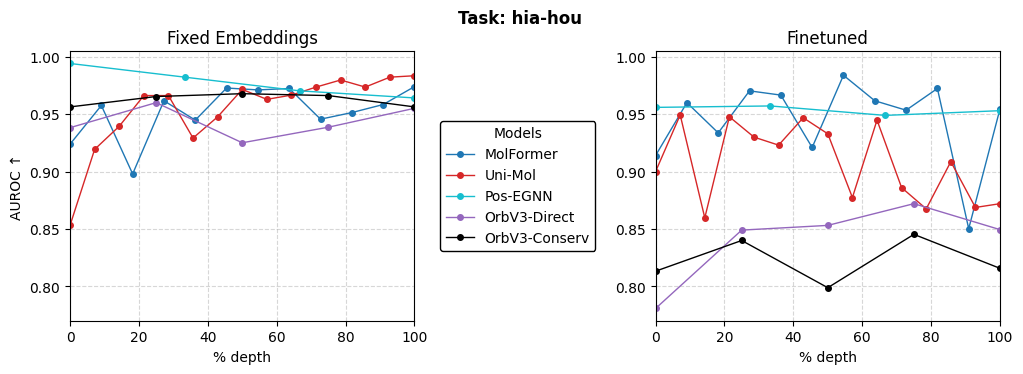}}
    \caption{AUROC (↑, the higher the better) on the \textbf{\textit{hia-hou}} dataset as a function of encoder depth (\% depth) for frozen embeddings (left) versus finetuned embeddings (right).}
\end{figure}

\begin{figure}[!htpb]
    \centering
    \makebox[\textwidth][c]{\includegraphics[width=1.0\linewidth]{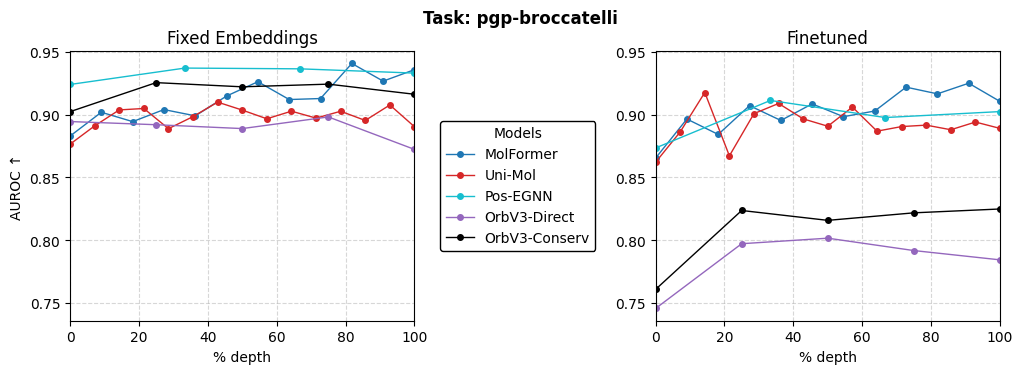}}
    \caption{AUROC (↑, the higher the better) on the \textbf{\textit{pgp-broccatelli}} dataset as a function of encoder depth (\% depth) for frozen embeddings (left) versus finetuned embeddings (right).}
\end{figure}

\begin{figure}[!htpb]
    \centering
    \makebox[\textwidth][c]{\includegraphics[width=1.0\linewidth]{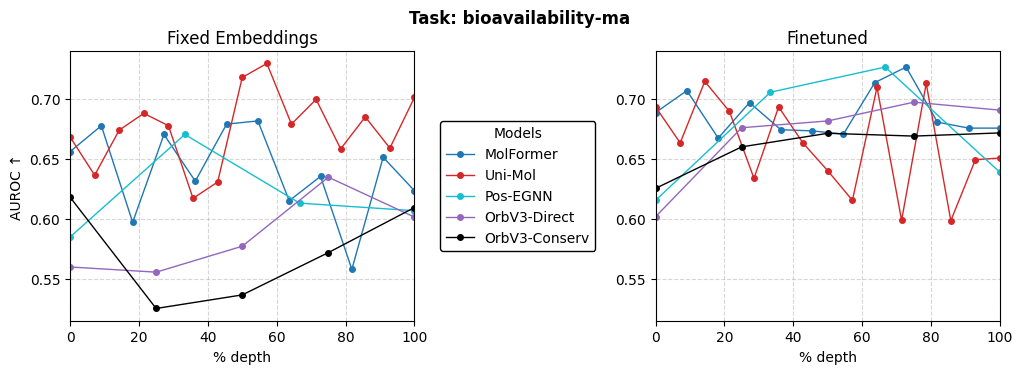}}
    \caption{AUROC (↑, the higher the better) on the \textbf{\textit{bioavailability-ma}} dataset as a function of encoder depth (\% depth) for frozen embeddings (left) versus finetuned embeddings (right).}
\end{figure}

\begin{figure}[!htpb]
    \centering
    \makebox[\textwidth][c]{\includegraphics[width=1.0\linewidth]{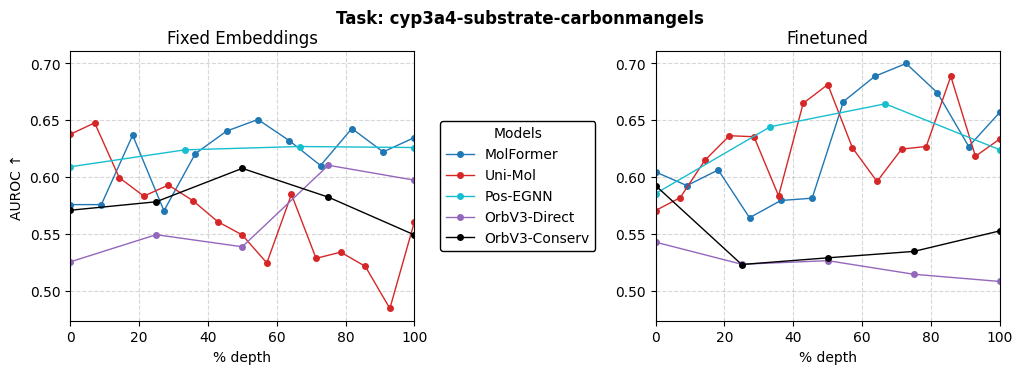}}
    \caption{AUROC (↑, the higher the better) on the \textbf{\textit{cyp3a4-substrate-carbonmangels}} dataset as a function of encoder depth (\% depth) for frozen embeddings (left) versus finetuned embeddings (right).}
\end{figure}

\begin{figure}[!htpb]
    \centering
    \makebox[\textwidth][c]{\includegraphics[width=1.0\linewidth]{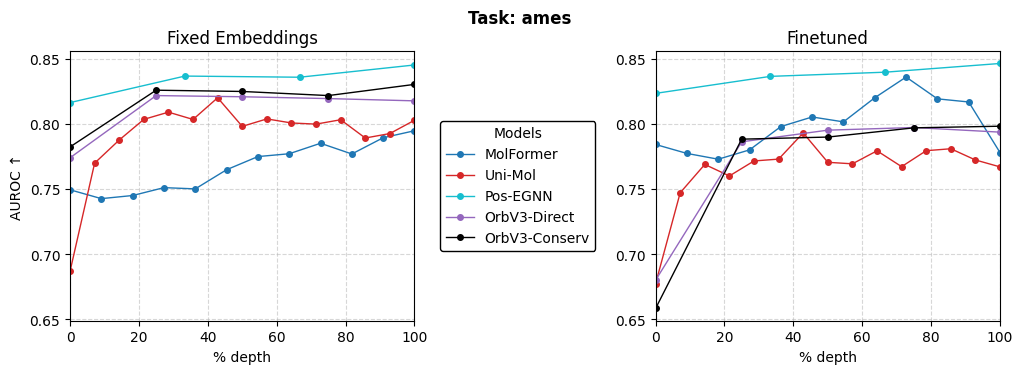}}
    \caption{AUROC (↑, the higher the better) on the \textbf{\textit{ames}} dataset as a function of encoder depth (\% depth) for frozen embeddings (left) versus finetuned embeddings (right).}
\end{figure}

\begin{figure}[!htpb]
    \centering
    \makebox[\textwidth][c]{\includegraphics[width=1.0\linewidth]{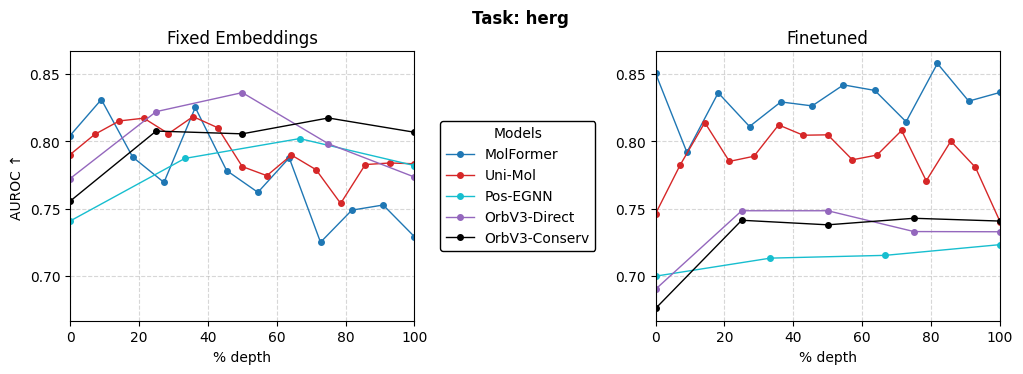}}
    \caption{AUROC (↑, the higher the better) on the \textbf{\textit{herg}} dataset as a function of encoder depth (\% depth) for frozen embeddings (left) versus finetuned embeddings (right).}
\end{figure}

\begin{figure}[!htpb]
    \centering
    \makebox[\textwidth][c]{\includegraphics[width=1.0\linewidth]{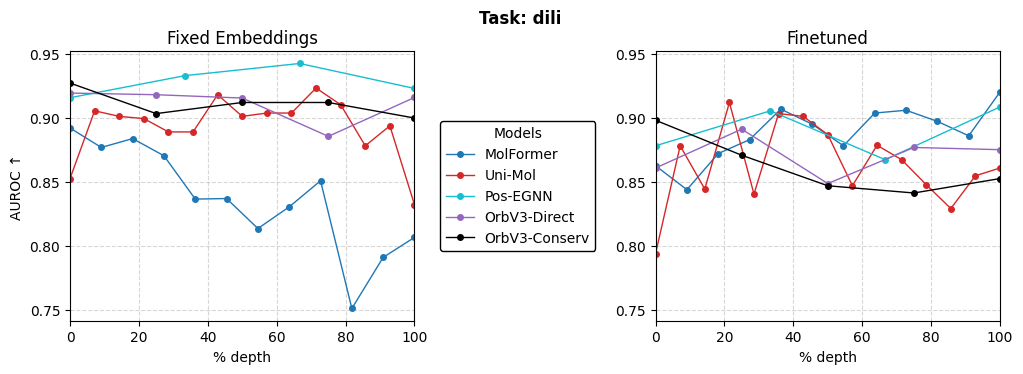}}
    \caption{AUROC (↑, the higher the better) on the \textbf{\textit{dili}} dataset as a function of encoder depth (\% depth) for frozen embeddings (left) versus finetuned embeddings (right).}
\end{figure}

\begin{figure}[!htpb]
    \centering
    \makebox[\textwidth][c]{\includegraphics[width=1.0\linewidth]{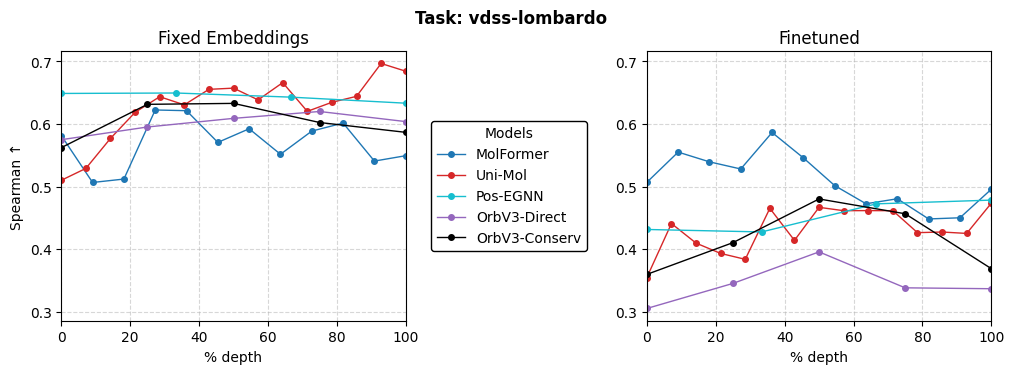}}
    \caption{SPEARMAN (↑, the higher the better) on the \textbf{\textit{vdss-lombardo}} dataset as a function of encoder depth (\% depth) for frozen embeddings (left) versus finetuned embeddings (right).}
\end{figure}

\begin{figure}[!htpb]
    \centering
    \makebox[\textwidth][c]{\includegraphics[width=1.0\linewidth]{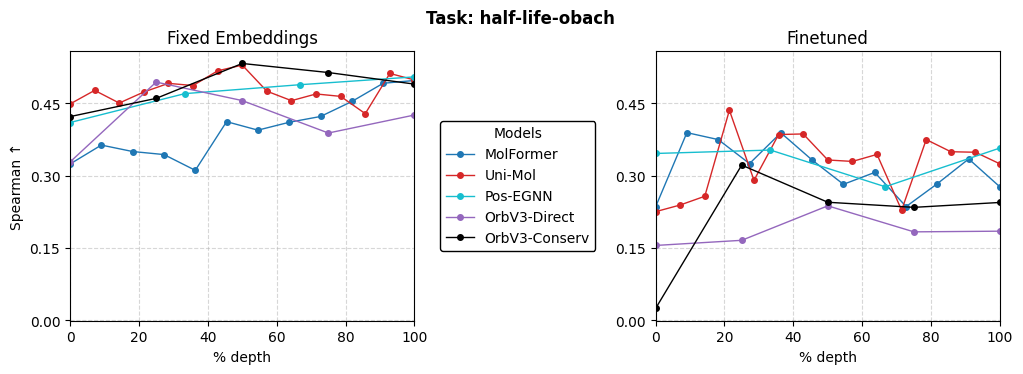}}
    \caption{SPEARMAN (↑, the higher the better) on the \textbf{\textit{half-life-obach}} dataset as a function of encoder depth (\% depth) for frozen embeddings (left) versus finetuned embeddings (right).}
\end{figure}

\begin{figure}[!htpb]
    \centering
    \makebox[\textwidth][c]{\includegraphics[width=1.0\linewidth]{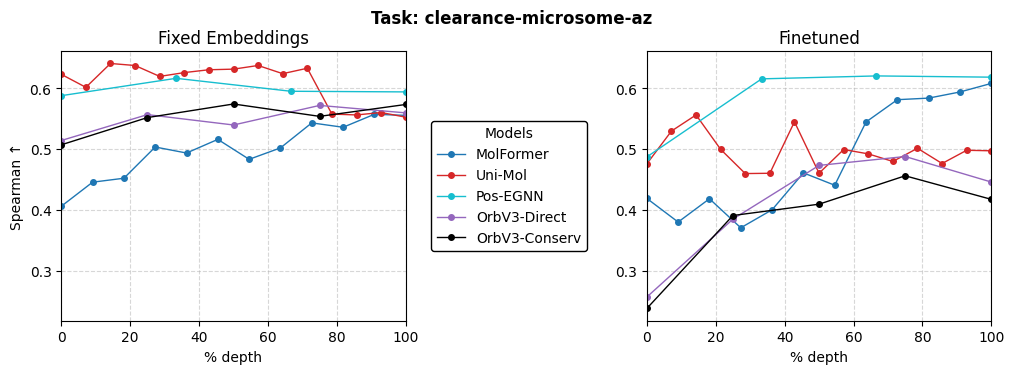}}
    \caption{SPEARMAN (↑, the higher the better) on the \textbf{\textit{clearance-microsome-az}} dataset as a function of encoder depth (\% depth) for frozen embeddings (left) versus finetuned embeddings (right).}
\end{figure}

\begin{figure}[!htpb]
    \centering
    \makebox[\textwidth][c]{\includegraphics[width=1.0\linewidth]{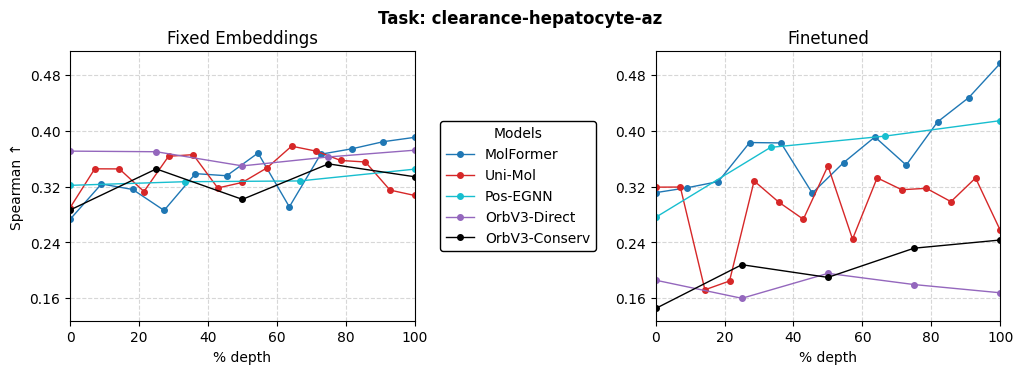}}
    \caption{SPEARMAN (↑, the higher the better) on the \textbf{\textit{clearance-hepatocyte-az}} dataset as a function of encoder depth (\% depth) for frozen embeddings (left) versus finetuned embeddings (right).}
\end{figure}

\begin{figure}[!htpb]
    \centering
    \makebox[\textwidth][c]{\includegraphics[width=1.0\linewidth]{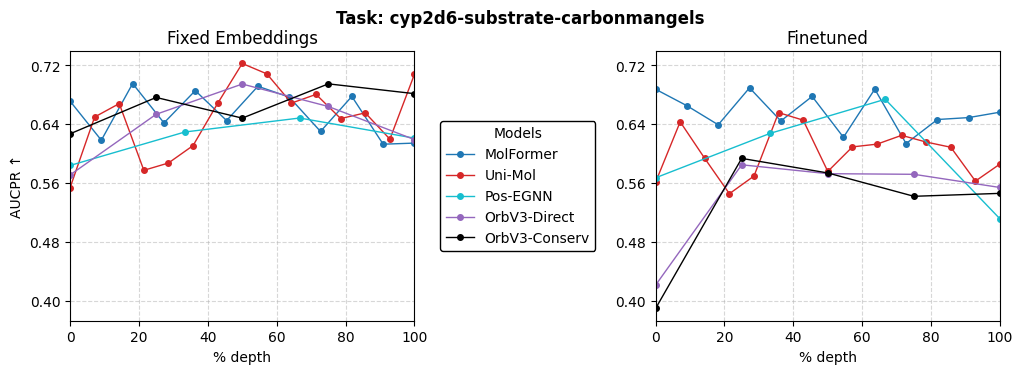}}
    \caption{SPEARMAN (↑, the higher the better) on the \textbf{\textit{cyp2d6-substrate-carbonmangels}} dataset as a function of encoder depth (\% depth) for frozen embeddings (left) versus finetuned embeddings (right).}
\end{figure}

\begin{figure}[!htpb]
    \centering
    \makebox[\textwidth][c]{\includegraphics[width=1.0\linewidth]{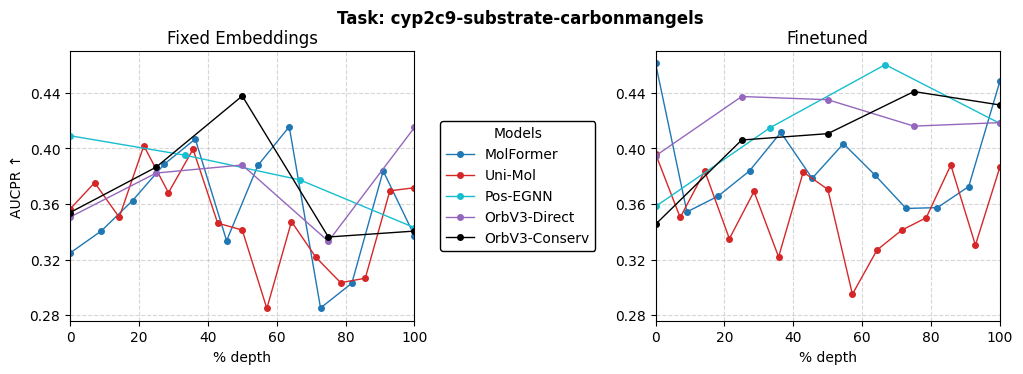}}
    \caption{AUCPR (↑, the higher the better) on the \textbf{\textit{cyp2c9-substrate-carbonmangels}} dataset as a function of encoder depth (\% depth) for frozen embeddings (left) versus finetuned embeddings (right).}
\end{figure}

\begin{figure}[!htpb]
    \centering
    \makebox[\textwidth][c]{\includegraphics[width=1.0\linewidth]{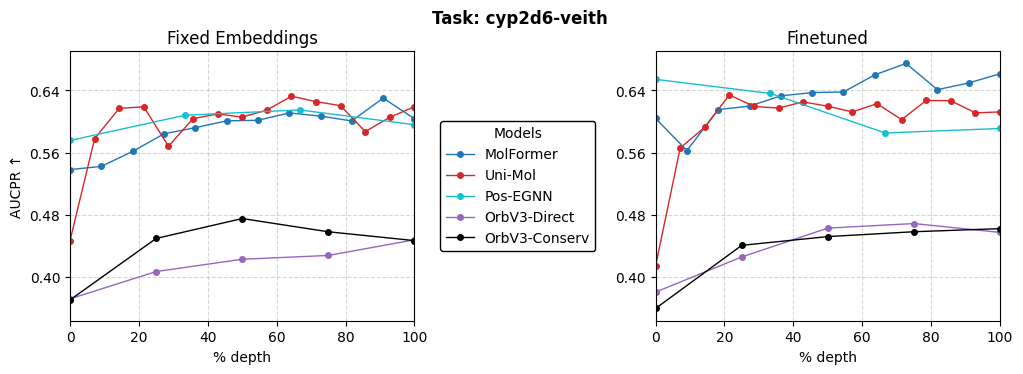}}
    \caption{AUCPR (↑, the higher the better) on the \textbf{\textit{cyp2d6-veith}} dataset as a function of encoder depth (\% depth) for frozen embeddings (left) versus finetuned embeddings (right).}
\end{figure}

\begin{figure}[!htpb]
    \centering
    \makebox[\textwidth][c]{\includegraphics[width=1.0\linewidth]{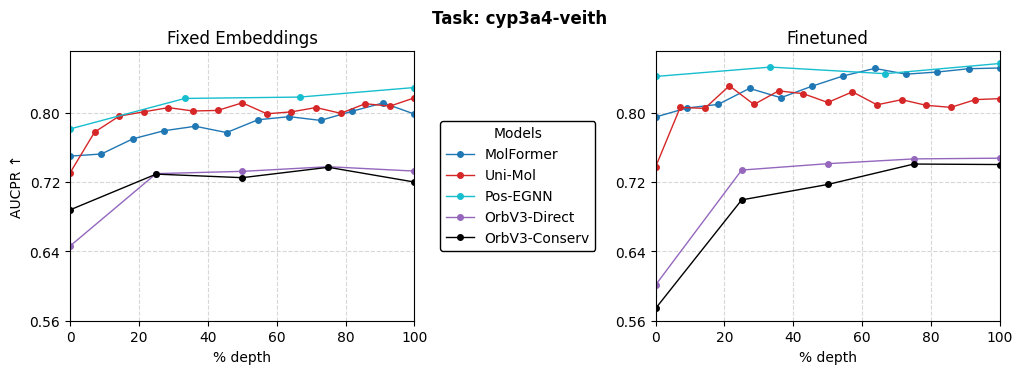}}
    \caption{AUCPR (↑, the higher the better) on the \textbf{\textit{cyp3a4-veith}} dataset as a function of encoder depth (\% depth) for frozen embeddings (left) versus finetuned embeddings (right).}
\end{figure}

\begin{figure}[!htpb]
    \centering
    \makebox[\textwidth][c]{\includegraphics[width=1.0\linewidth]{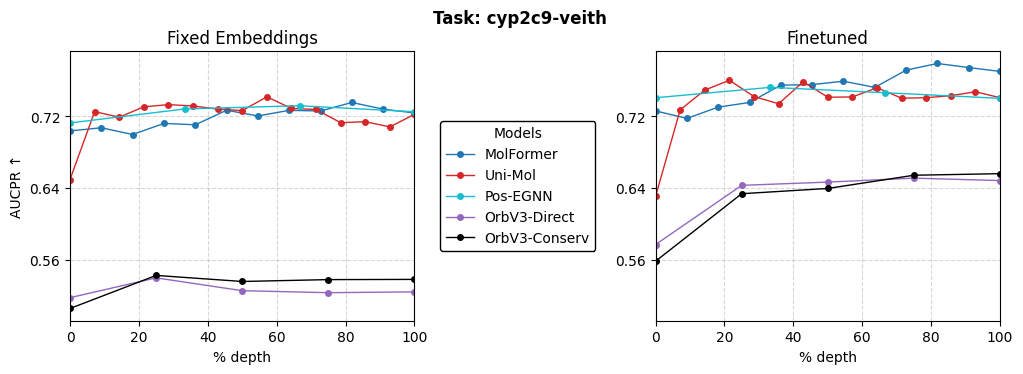}}
    \caption{AUCPR (↑, the higher the better) on the \textbf{\textit{cyp2c9-veith}} dataset as a function of encoder depth (\% depth) for frozen embeddings (left) versus finetuned embeddings (right).}
\end{figure}

\newpage

\section{MolFormer: Frozen Embedding vs. Finetuned Scatter Plots} \label{appendix D}

This appendix section presents a series of scatter plots dedicated to the MolFormer model. Each plot illustrates the relationship between frozen embedding performance and full finetuning performance across the model's different layers for a specific downstream task. The plots were ordered by Pearson correlation.


\begin{figure}[!htpb]
  \centering
  \begin{subfigure}[b]{0.49\textwidth}
    \centering
    \includegraphics[width=\textwidth]{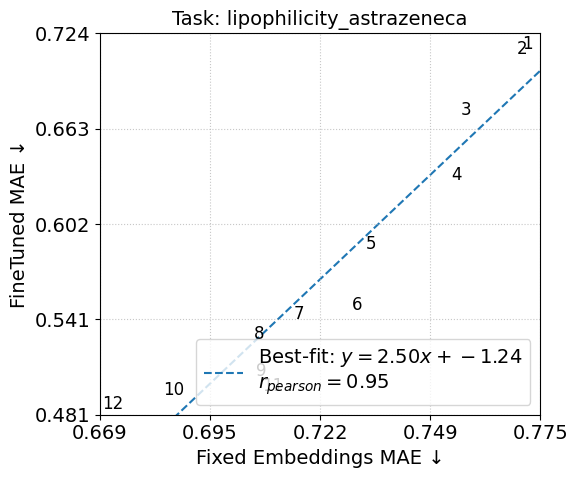}
  \end{subfigure}
  \hfill
  \begin{subfigure}[b]{0.49\textwidth}
    \centering
    \includegraphics[width=\textwidth]{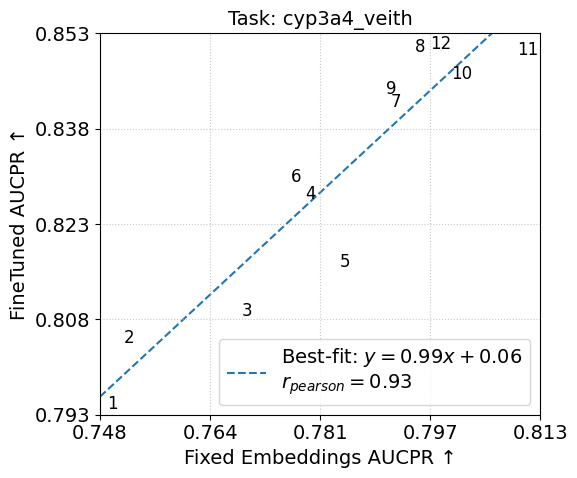}
  \end{subfigure}
  \caption{Frozen embeddings vs. finetuned performance for every layer of MolFormer.}
\end{figure}

\begin{figure}[!htpb]
  \centering
  \begin{subfigure}[b]{0.49\textwidth}
    \centering
    \includegraphics[width=\textwidth]{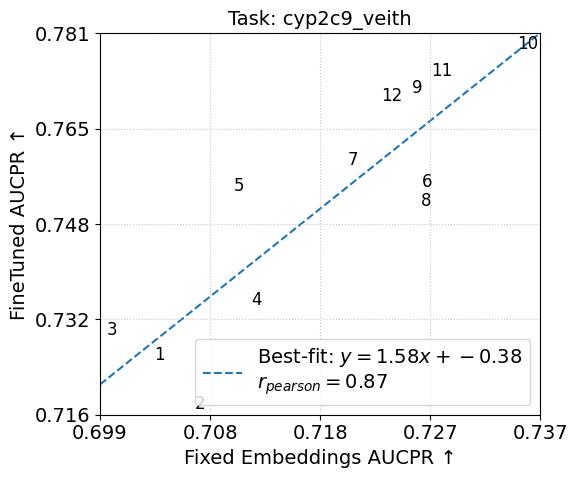}
  \end{subfigure}
  \hfill
  \begin{subfigure}[b]{0.49\textwidth}
    \centering
    \includegraphics[width=\textwidth]{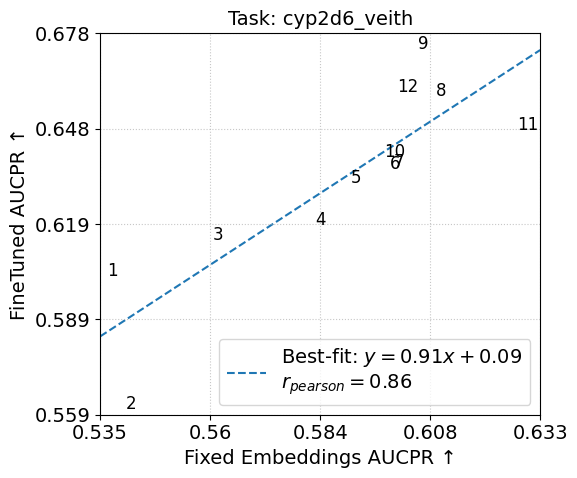}
  \end{subfigure}
  \caption{Frozen embeddings vs. finetuned performance for every layer of MolFormer.}
\end{figure}

\begin{figure}[!htpb]
  \centering
  \begin{subfigure}[b]{0.49\textwidth}
    \centering
    \includegraphics[width=\textwidth]{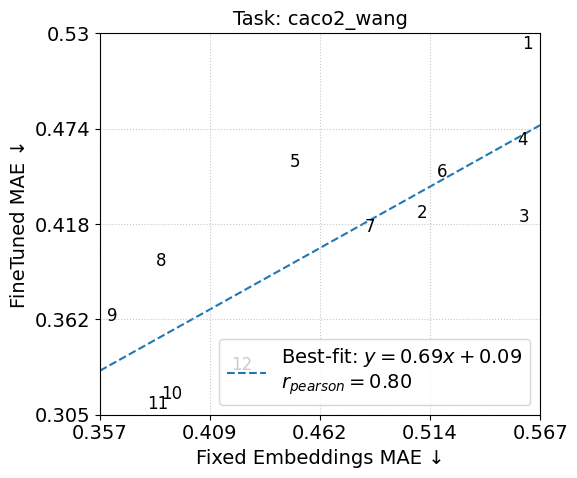}
  \end{subfigure}
  \hfill
  \begin{subfigure}[b]{0.49\textwidth}
    \centering
    \includegraphics[width=\textwidth]{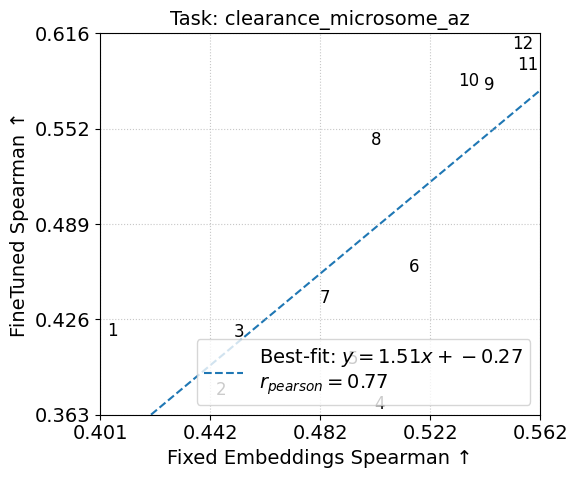}
  \end{subfigure}
  \caption{Frozen embeddings vs. finetuned performance for every layer of MolFormer.}
\end{figure}

\begin{figure}[!htpb]
  \centering
  \begin{subfigure}[b]{0.49\textwidth}
    \centering
    \includegraphics[width=\textwidth]{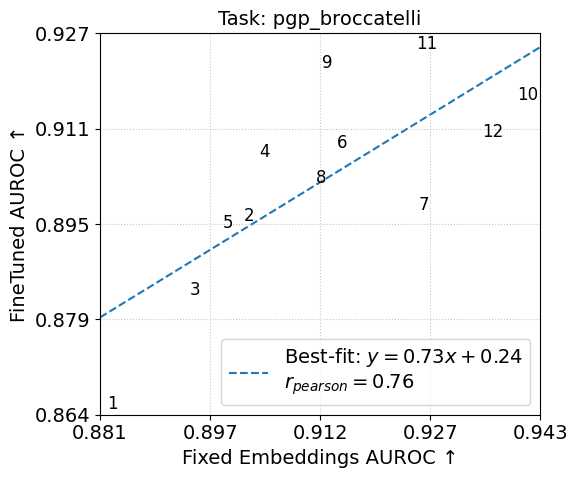}
  \end{subfigure}
  \hfill
  \begin{subfigure}[b]{0.49\textwidth}
    \centering
    \includegraphics[width=\textwidth]{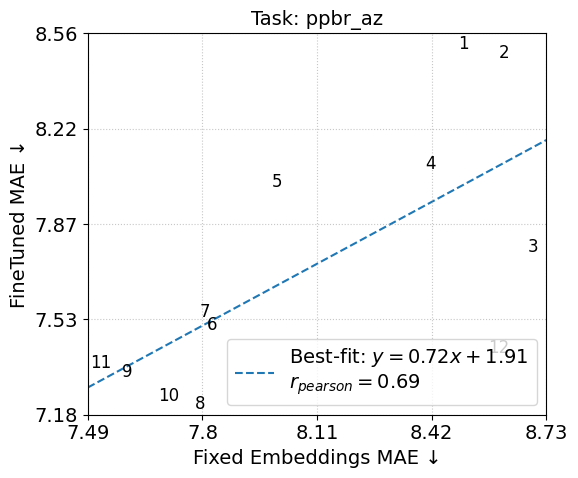}
  \end{subfigure}
  \caption{Frozen embeddings vs. finetuned performance for every layer of MolFormer.}
\end{figure}

\begin{figure}[!htpb]
  \centering
  \begin{subfigure}[b]{0.49\textwidth}
    \centering
    \includegraphics[width=\textwidth]{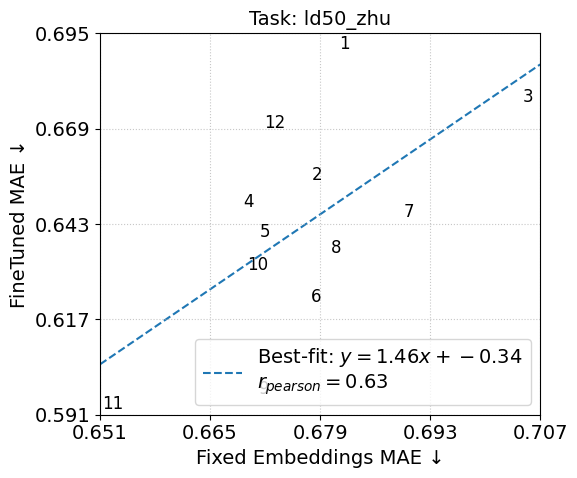}
  \end{subfigure}
  \hfill
  \begin{subfigure}[b]{0.49\textwidth}
    \centering
    \includegraphics[width=\textwidth]{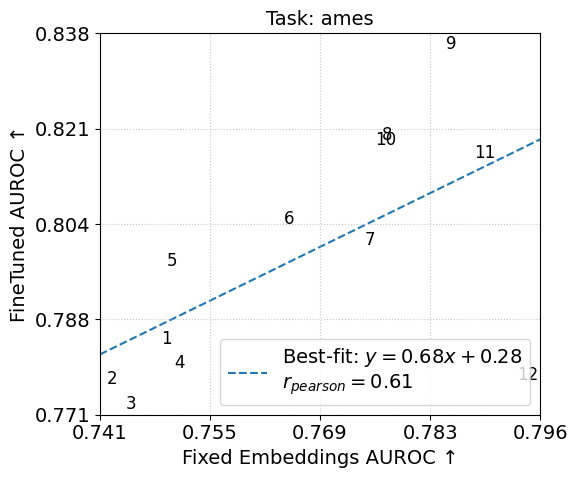}
  \end{subfigure}
  \caption{Frozen embeddings vs. finetuned performance for every layer of MolFormer.}
\end{figure}

\begin{figure}[!htpb]
  \centering
  \begin{subfigure}[b]{0.49\textwidth}
    \centering
    \includegraphics[width=\textwidth]{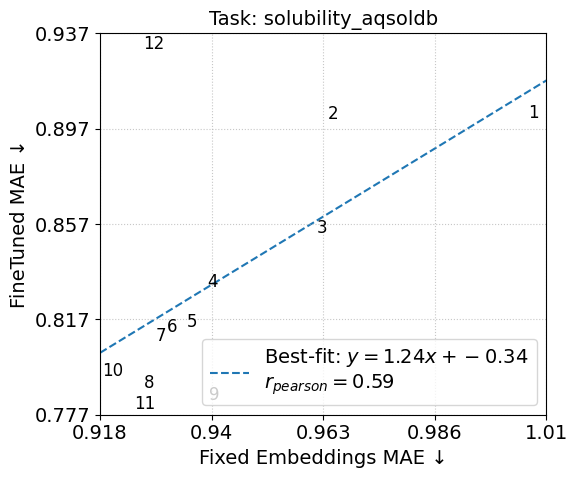}
  \end{subfigure}
  \hfill
  \begin{subfigure}[b]{0.49\textwidth}
    \centering
    \includegraphics[width=\textwidth]{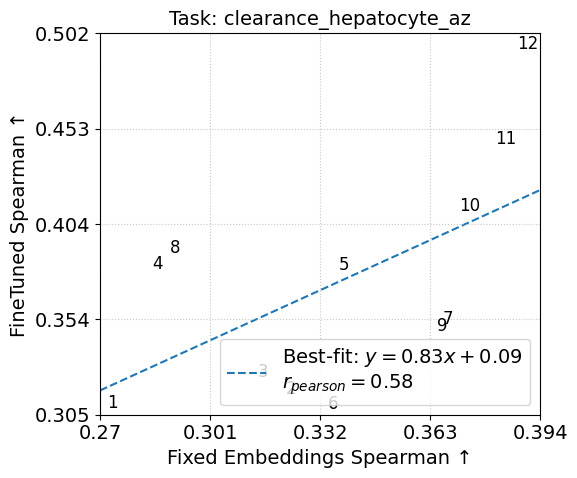}
  \end{subfigure}
  \caption{Frozen embeddings vs. finetuned performance for every layer of MolFormer.}
\end{figure}

\begin{figure}[!htpb]
  \centering
  \begin{subfigure}[b]{0.49\textwidth}
    \centering
    \includegraphics[width=\textwidth]{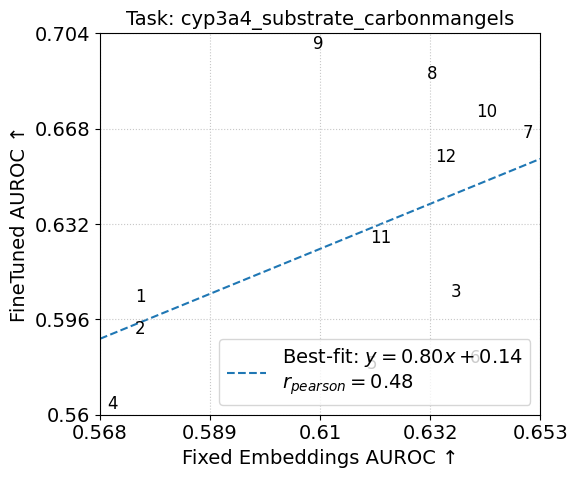}
  \end{subfigure}
  \hfill
  \begin{subfigure}[b]{0.49\textwidth}
    \centering
    \includegraphics[width=\textwidth]{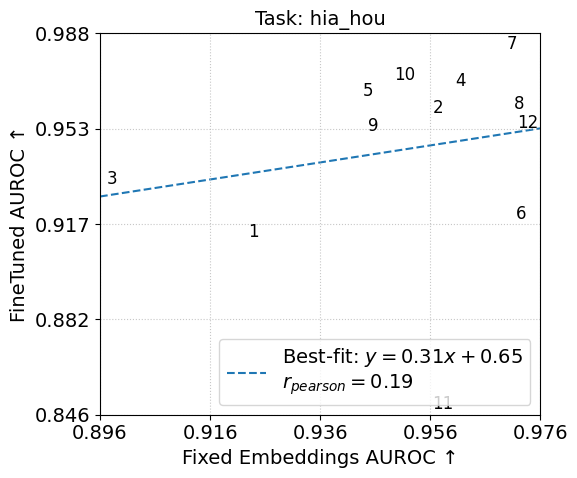}
  \end{subfigure}
  \caption{Frozen embeddings vs. finetuned performance for every layer of MolFormer.}
\end{figure}

\begin{figure}[!htpb]
  \centering
  \begin{subfigure}[b]{0.49\textwidth}
    \centering
    \includegraphics[width=\textwidth]{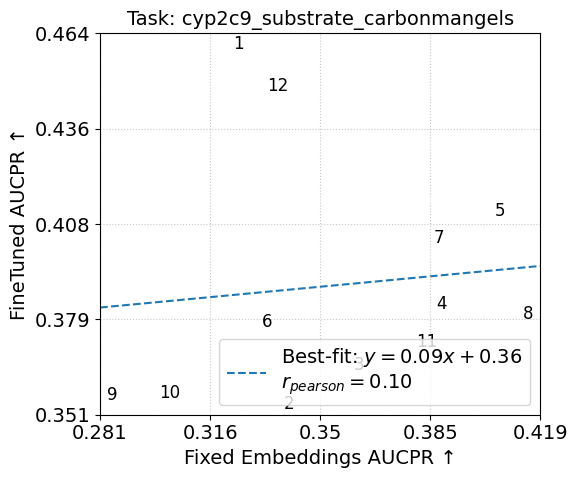}
  \end{subfigure}
  \hfill
  \begin{subfigure}[b]{0.49\textwidth}
    \centering
    \includegraphics[width=\textwidth]{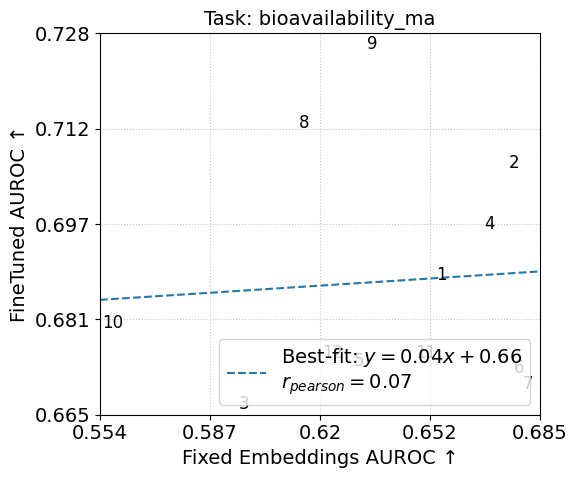}
  \end{subfigure}
  \caption{Frozen embeddings vs. finetuned performance for every layer of MolFormer.}
\end{figure}

\begin{figure}[!htpb]
  \centering
  \begin{subfigure}[b]{0.49\textwidth}
    \centering
    \includegraphics[width=\textwidth]{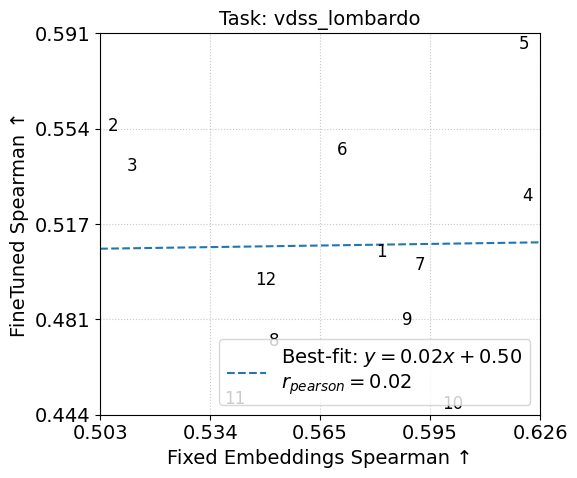}
  \end{subfigure}
  \hfill
  \begin{subfigure}[b]{0.49\textwidth}
    \centering
    \includegraphics[width=\textwidth]{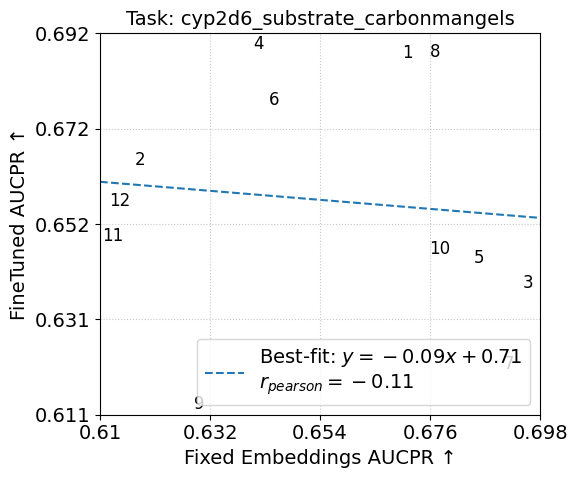}
  \end{subfigure}
  \caption{Frozen embeddings vs. finetuned performance for every layer of MolFormer.}
\end{figure}

\begin{figure}[!htpb]
  \centering
  \begin{subfigure}[b]{0.49\textwidth}
    \centering
    \includegraphics[width=\textwidth]{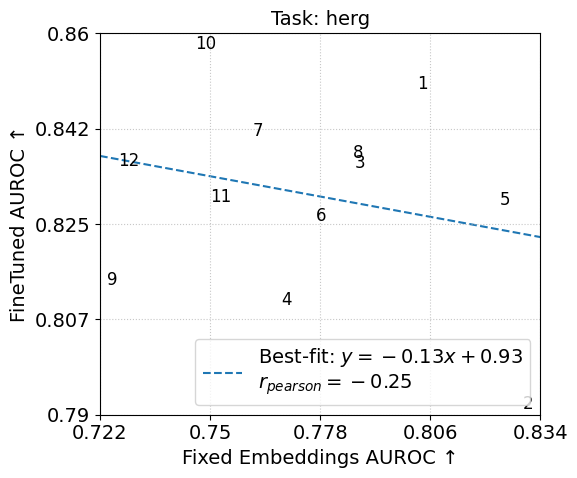}
  \end{subfigure}
  \hfill
  \begin{subfigure}[b]{0.49\textwidth}
    \centering
    \includegraphics[width=\textwidth]{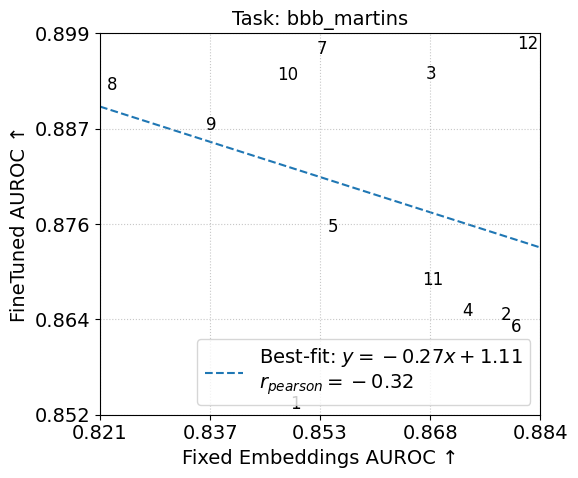}
  \end{subfigure}
  \caption{Frozen embeddings vs. finetuned performance for every layer of MolFormer.}
\end{figure}

\begin{figure}[!htpb]
  \centering
  \begin{subfigure}[b]{0.49\textwidth}
    \centering
    \includegraphics[width=\textwidth]{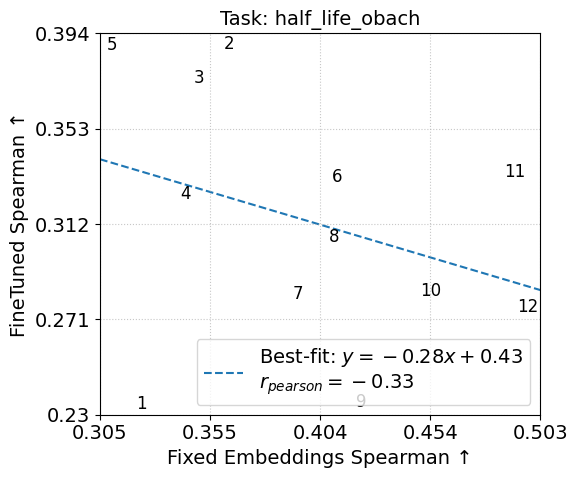}
  \end{subfigure}
  \hfill
  \begin{subfigure}[b]{0.49\textwidth}
    \centering
    \includegraphics[width=\textwidth]{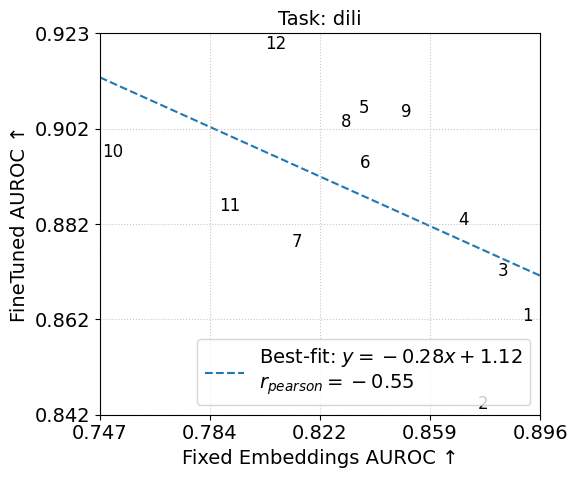}
  \end{subfigure}
  \caption{Frozen embeddings vs. finetuned performance for every layer of MolFormer.}
\end{figure}

\section{Uni-Mol: Frozen embedding vs. Finetuned Scatter Plots} \label{appendix E}

This appendix section presents a series of scatter plots dedicated to the Uni-Mol model. Each plot illustrates the relationship between frozen embedding performance and full finetuning performance across the model's different layers for a specific downstream task. The plots were ordered by Pearson correlation.

\begin{figure}[!htpb]
  \centering
  \begin{subfigure}[b]{0.49\textwidth}
    \centering
    \includegraphics[width=\textwidth]{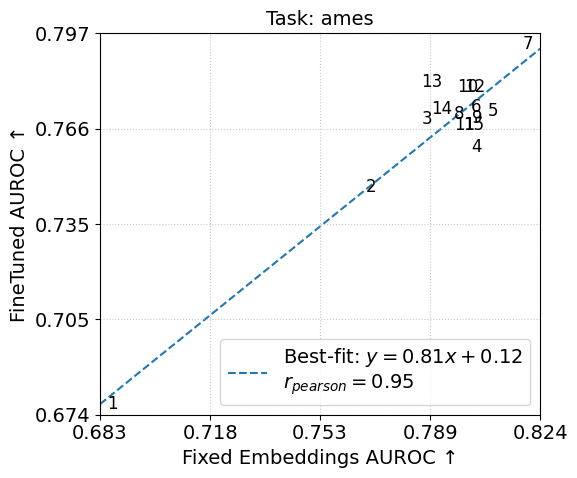}
  \end{subfigure}
  \hfill
  \begin{subfigure}[b]{0.49\textwidth}
    \centering
    \includegraphics[width=\textwidth]{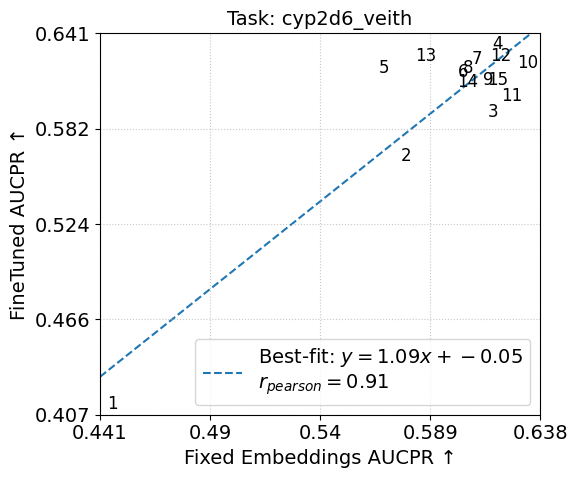}
  \end{subfigure}
  \caption{Frozen embeddings vs. finetuned performance for every layer of Uni-Mol.}
\end{figure}

\begin{figure}[!htpb]
  \centering
  \begin{subfigure}[b]{0.49\textwidth}
    \centering
    \includegraphics[width=\textwidth]{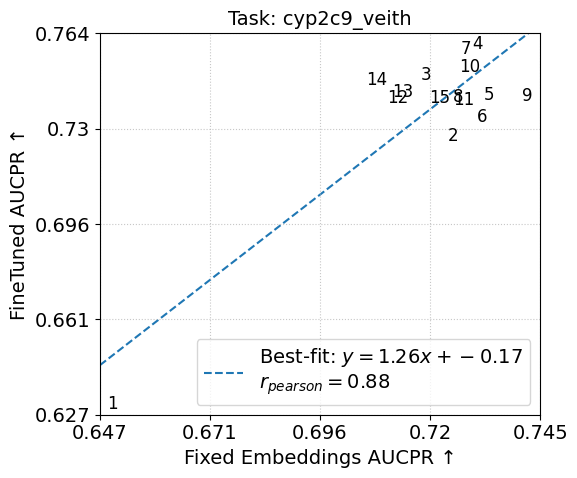}
  \end{subfigure}
  \hfill
  \begin{subfigure}[b]{0.49\textwidth}
    \centering
    \includegraphics[width=\textwidth]{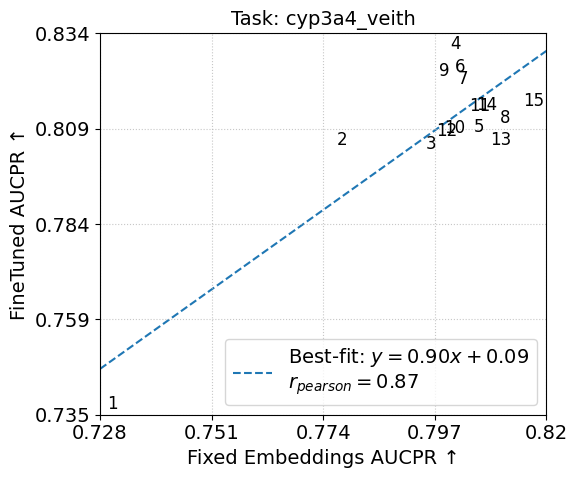}
  \end{subfigure}
  \caption{Frozen embeddings vs. finetuned performance for every layer of Uni-Mol.}
\end{figure}

\begin{figure}[!htpb]
  \centering
  \begin{subfigure}[b]{0.49\textwidth}
    \centering
    \includegraphics[width=\textwidth]{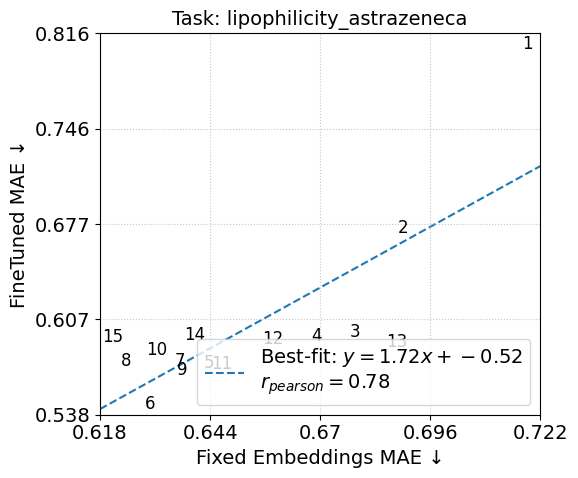}
  \end{subfigure}
  \hfill
  \begin{subfigure}[b]{0.49\textwidth}
    \centering
    \includegraphics[width=\textwidth]{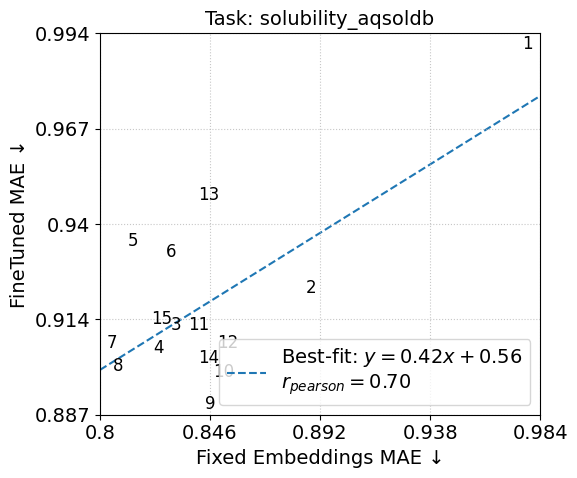}
  \end{subfigure}
  \caption{Frozen embeddings vs. finetuned performance for every layer of Uni-Mol.}
\end{figure}

\begin{figure}[!htpb]
  \centering
  \begin{subfigure}[b]{0.49\textwidth}
    \centering
    \includegraphics[width=\textwidth]{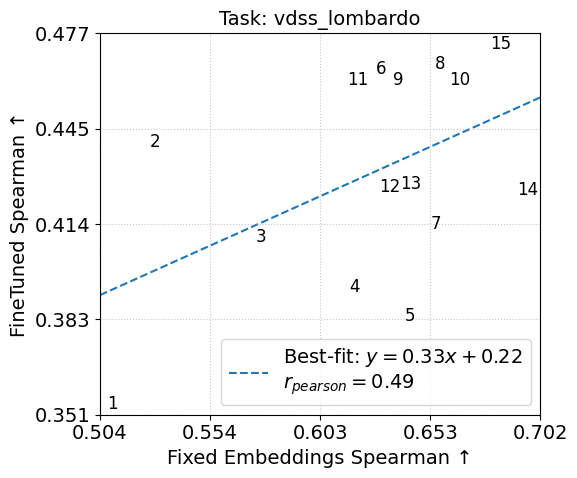}
  \end{subfigure}
  \hfill
  \begin{subfigure}[b]{0.49\textwidth}
    \centering
    \includegraphics[width=\textwidth]{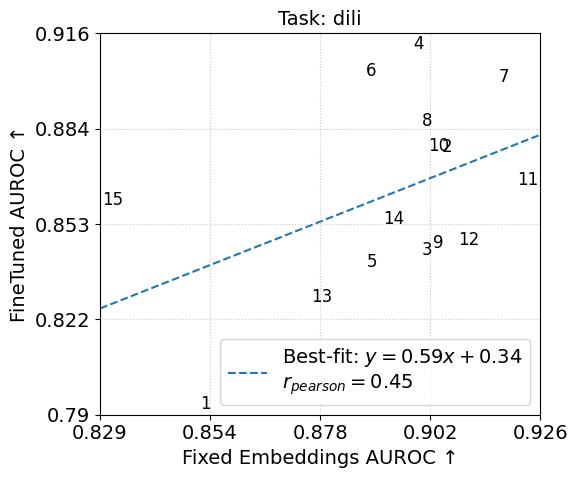}
  \end{subfigure}
  \caption{Frozen embeddings vs. finetuned performance for every layer of Uni-Mol.}
\end{figure}

\begin{figure}[!htpb]
  \centering
  \begin{subfigure}[b]{0.49\textwidth}
    \centering
    \includegraphics[width=\textwidth]{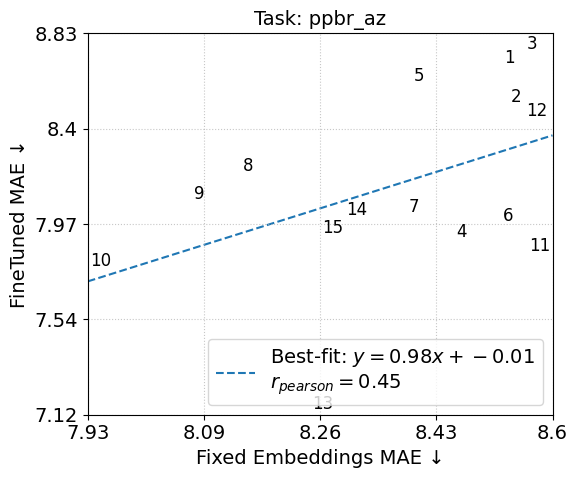}
  \end{subfigure}
  \hfill
  \begin{subfigure}[b]{0.49\textwidth}
    \centering
    \includegraphics[width=\textwidth]{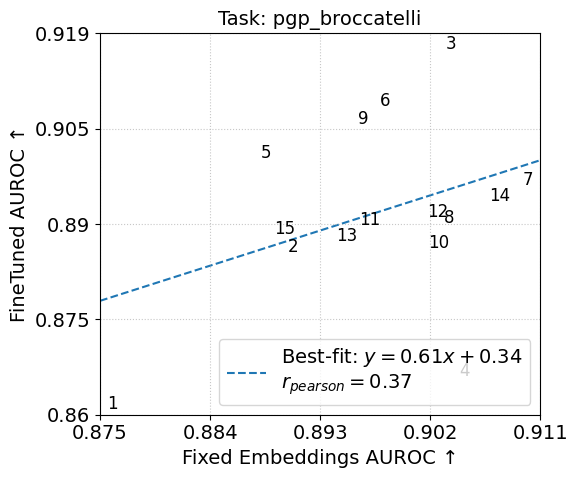}
  \end{subfigure}
  \caption{Frozen embeddings vs. finetuned performance for every layer of Uni-Mol.}
\end{figure}

\begin{figure}[!htpb]
  \centering
  \begin{subfigure}[b]{0.49\textwidth}
    \centering
    \includegraphics[width=\textwidth]{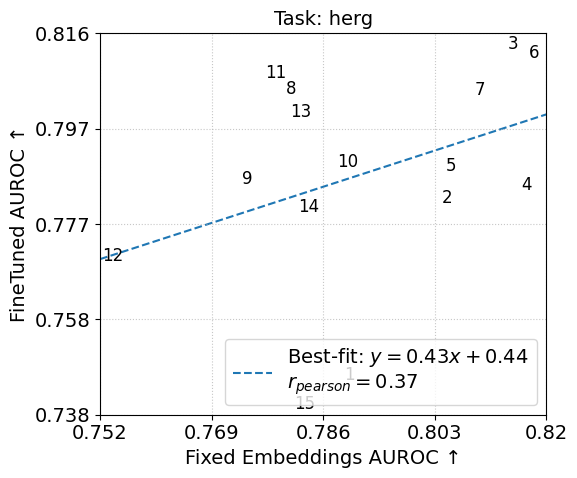}
  \end{subfigure}
  \hfill
  \begin{subfigure}[b]{0.49\textwidth}
    \centering
    \includegraphics[width=\textwidth]{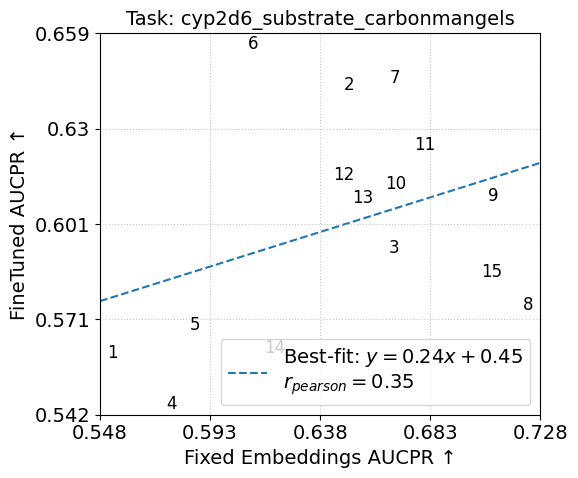}
  \end{subfigure}
  \caption{Frozen embeddings vs. finetuned performance for every layer of Uni-Mol.}
\end{figure}

\begin{figure}[!htpb]
  \centering
  \begin{subfigure}[b]{0.49\textwidth}
    \centering
    \includegraphics[width=\textwidth]{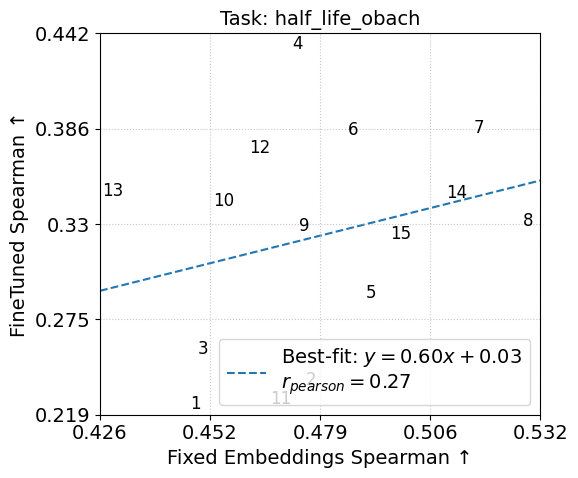}
  \end{subfigure}
  \hfill
  \begin{subfigure}[b]{0.49\textwidth}
    \centering
    \includegraphics[width=\textwidth]{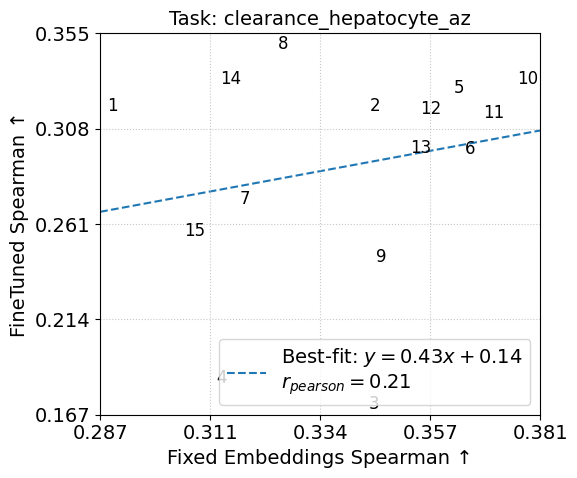}
  \end{subfigure}
  \caption{Frozen embeddings vs. finetuned performance for every layer of Uni-Mol.}
\end{figure}

\begin{figure}[!htpb]
  \centering
  \begin{subfigure}[b]{0.49\textwidth}
    \centering
    \includegraphics[width=\textwidth]{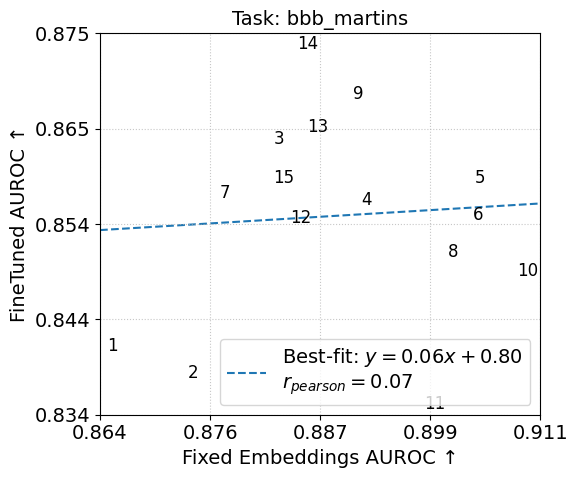}
  \end{subfigure}
  \hfill
  \begin{subfigure}[b]{0.49\textwidth}
    \centering
    \includegraphics[width=\textwidth]{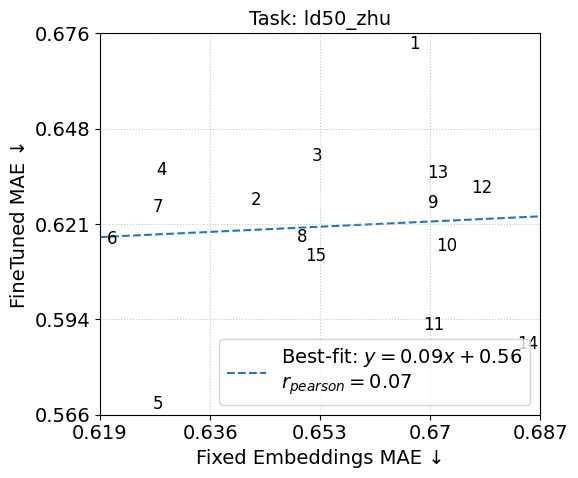}
  \end{subfigure}
  \caption{Frozen embeddings vs. finetuned performance for every layer of Uni-Mol.}
\end{figure}

\begin{figure}[!htpb]
  \centering
  \begin{subfigure}[b]{0.49\textwidth}
    \centering
    \includegraphics[width=\textwidth]{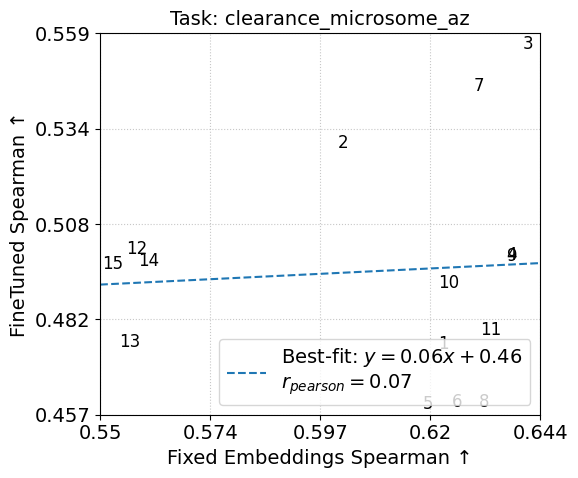}
  \end{subfigure}
  \hfill
  \begin{subfigure}[b]{0.49\textwidth}
    \centering
    \includegraphics[width=\textwidth]{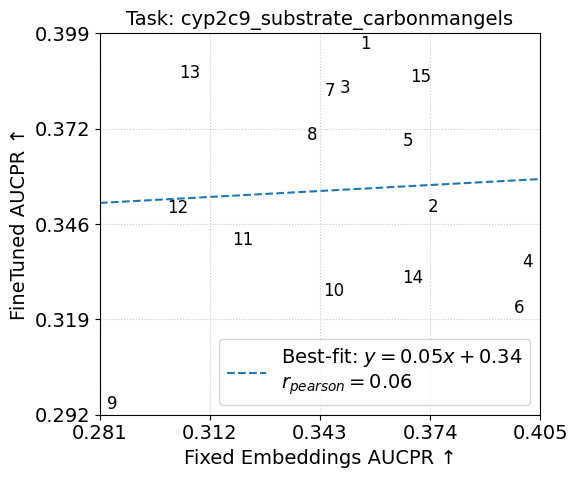}
  \end{subfigure}
  \caption{Frozen embeddings vs. finetuned performance for every layer of Uni-Mol.}
\end{figure}

\begin{figure}[!htpb]
  \centering
  \begin{subfigure}[b]{0.49\textwidth}
    \centering
    \includegraphics[width=\textwidth]{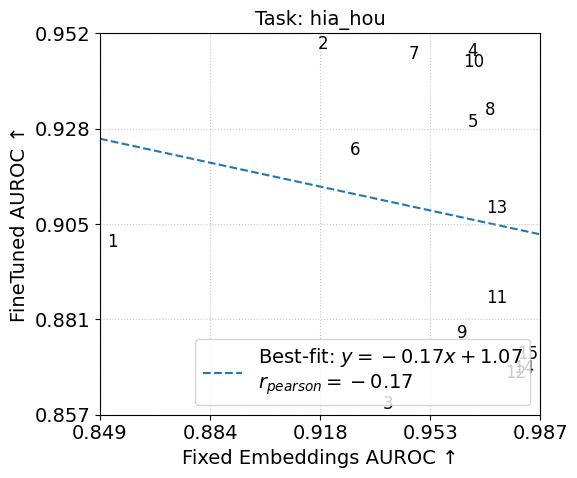}
  \end{subfigure}
  \hfill
  \begin{subfigure}[b]{0.49\textwidth}
    \centering
    \includegraphics[width=\textwidth]{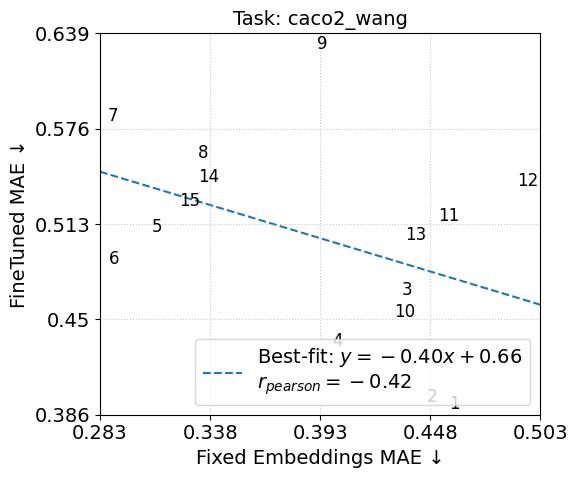}
  \end{subfigure}
  \caption{Frozen embeddings vs. finetuned performance for every layer of Uni-Mol.}
\end{figure}

\begin{figure}[!htpb]
  \centering
  \begin{subfigure}[b]{0.49\textwidth}
    \centering
    \includegraphics[width=\textwidth]{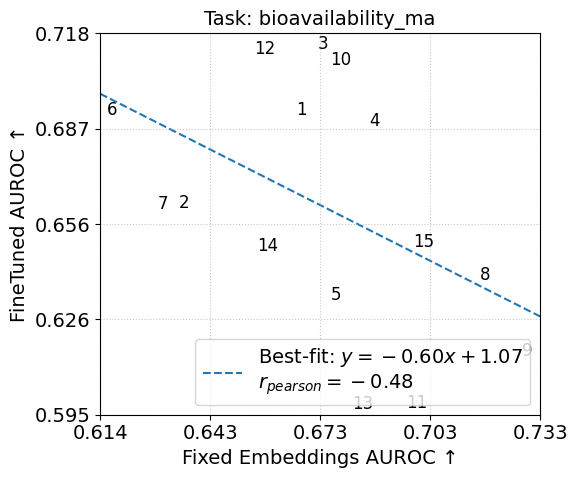}
  \end{subfigure}
  \hfill
  \begin{subfigure}[b]{0.49\textwidth}
    \centering
    \includegraphics[width=\textwidth]{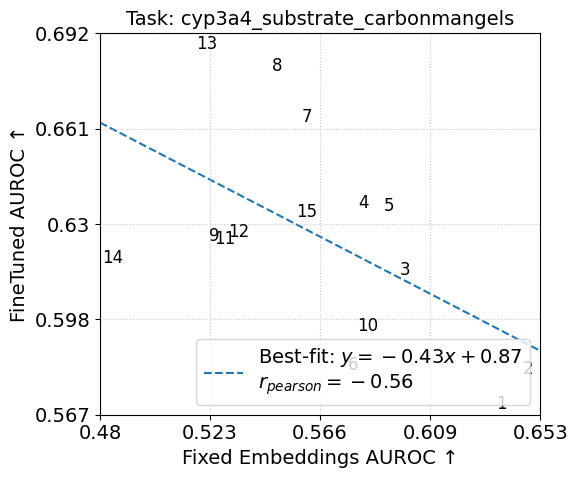}
  \end{subfigure}
  \caption{Frozen embeddings vs. finetuned performance for every layer of Uni-Mol.}
\end{figure}

\newpage

\section{OrbV3-Direct: Frozen embedding vs. Finetuned Scatter Plots} \label{appendix F}

This appendix section presents a series of scatter plots dedicated to the OrbV3-Direct model. Each plot illustrates the relationship between frozen embedding performance and full finetuning performance across the model's different layers for a specific downstream task. The plots were ordered by Pearson correlation.

\begin{figure}[!htpb]
  \centering
  \begin{subfigure}[b]{0.49\textwidth}
    \centering
    \includegraphics[width=\textwidth]{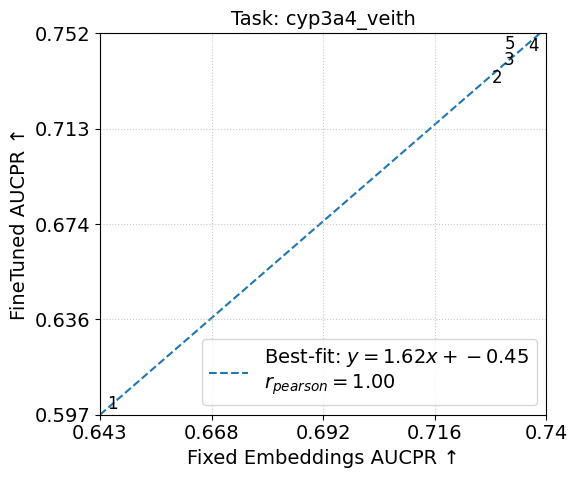}
  \end{subfigure}
  \hfill
  \begin{subfigure}[b]{0.49\textwidth}
    \centering
    \includegraphics[width=\textwidth]{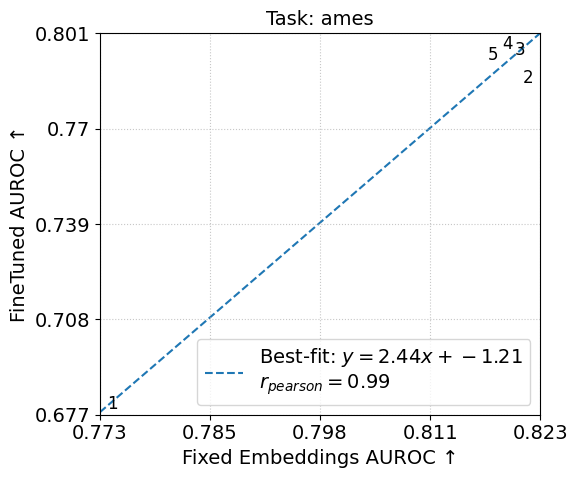}
  \end{subfigure}
  \caption{Frozen embeddings vs. finetuned performance for every layer of OrbV3-Direct.}
\end{figure}

\begin{figure}[!htpb]
  \centering
  \begin{subfigure}[b]{0.49\textwidth}
    \centering
    \includegraphics[width=\textwidth]{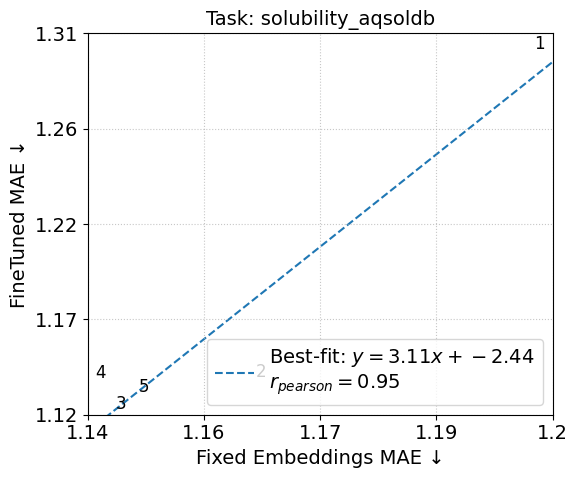}
  \end{subfigure}
  \hfill
  \begin{subfigure}[b]{0.49\textwidth}
    \centering
    \includegraphics[width=\textwidth]{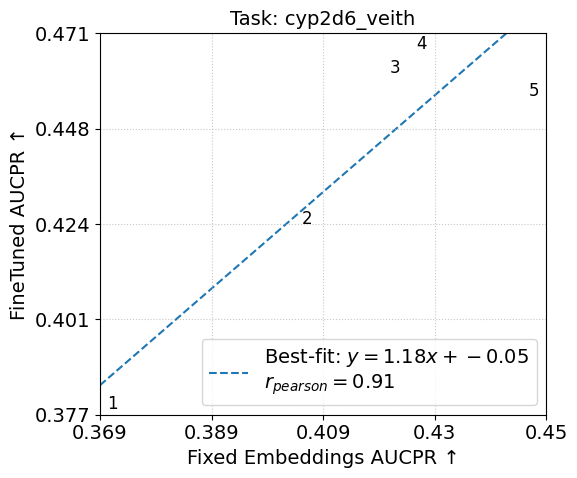}
  \end{subfigure}
  \caption{Frozen embeddings vs. finetuned performance for every layer of OrbV3-Direct.}
\end{figure}

\begin{figure}[!htpb]
  \centering
  \begin{subfigure}[b]{0.49\textwidth}
    \centering
    \includegraphics[width=\textwidth]{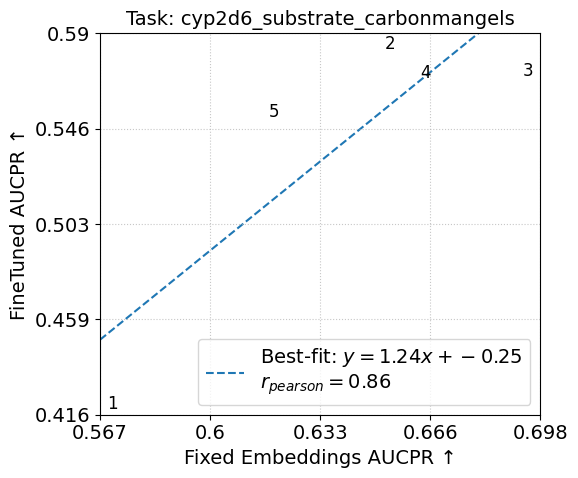}
  \end{subfigure}
  \hfill
  \begin{subfigure}[b]{0.49\textwidth}
    \centering
    \includegraphics[width=\textwidth]{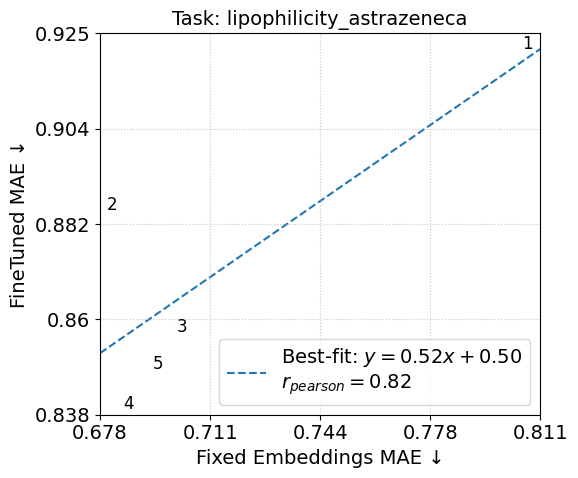}
  \end{subfigure}
  \caption{Frozen embeddings vs. finetuned performance for every layer of OrbV3-Direct.}
\end{figure}

\begin{figure}[!htpb]
  \centering
  \begin{subfigure}[b]{0.49\textwidth}
    \centering
    \includegraphics[width=\textwidth]{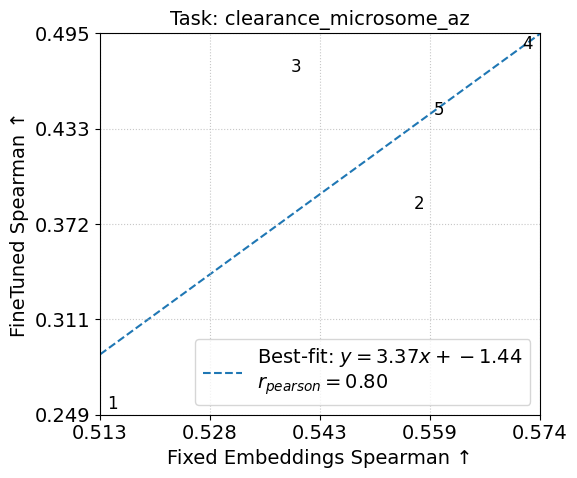}
  \end{subfigure}
  \hfill
  \begin{subfigure}[b]{0.49\textwidth}
    \centering
    \includegraphics[width=\textwidth]{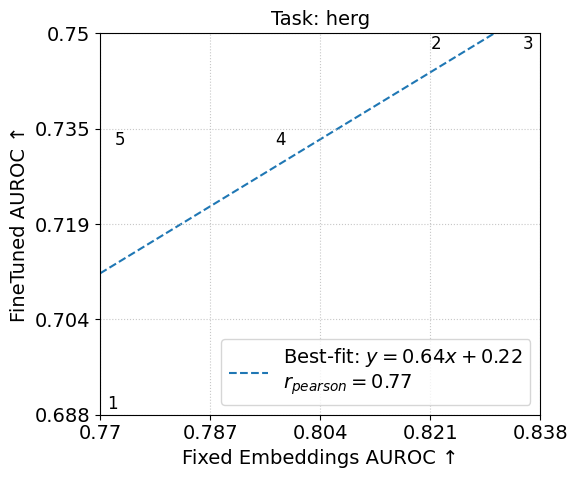}
  \end{subfigure}
  \caption{Frozen embeddings vs. finetuned performance for every layer of OrbV3-Direct.}
\end{figure}

\begin{figure}[!htpb]
  \centering
  \begin{subfigure}[b]{0.49\textwidth}
    \centering
    \includegraphics[width=\textwidth]{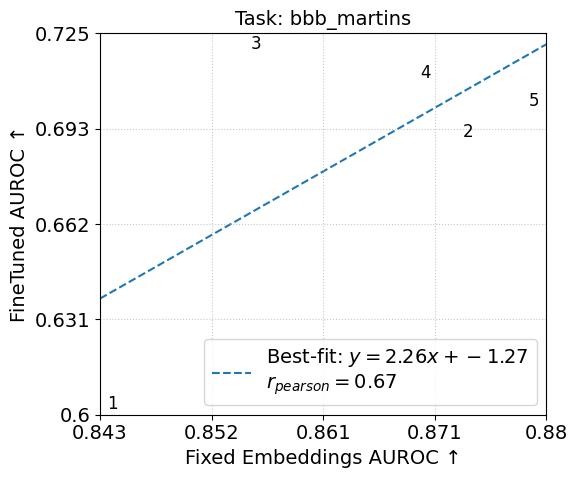}
  \end{subfigure}
  \hfill
  \begin{subfigure}[b]{0.49\textwidth}
    \centering
    \includegraphics[width=\textwidth]{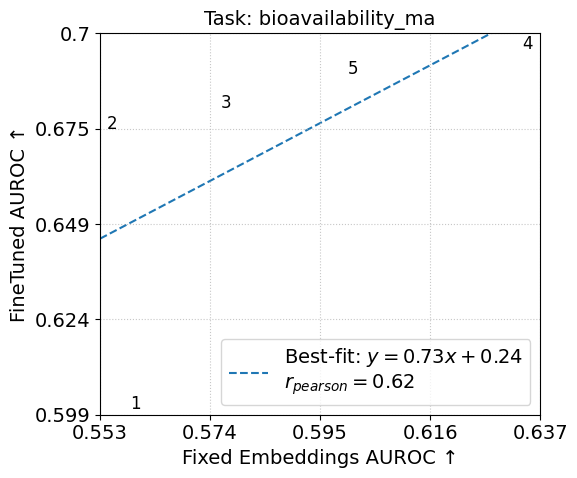}
  \end{subfigure}
  \caption{Frozen embeddings vs. finetuned performance for every layer of OrbV3-Direct.}
\end{figure}

\begin{figure}[!htpb]
  \centering
  \begin{subfigure}[b]{0.49\textwidth}
    \centering
    \includegraphics[width=\textwidth]{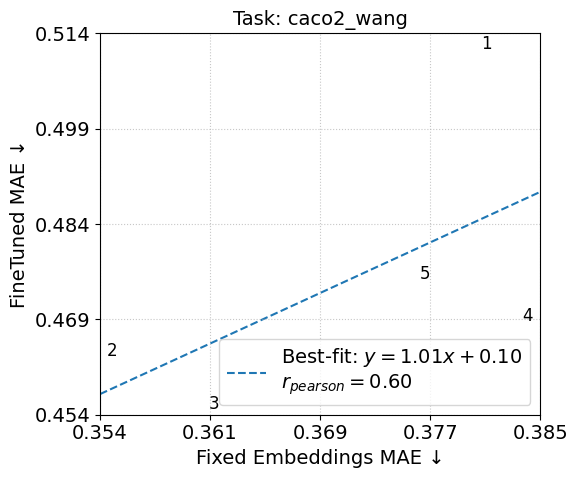}
  \end{subfigure}
  \hfill
  \begin{subfigure}[b]{0.49\textwidth}
    \centering
    \includegraphics[width=\textwidth]{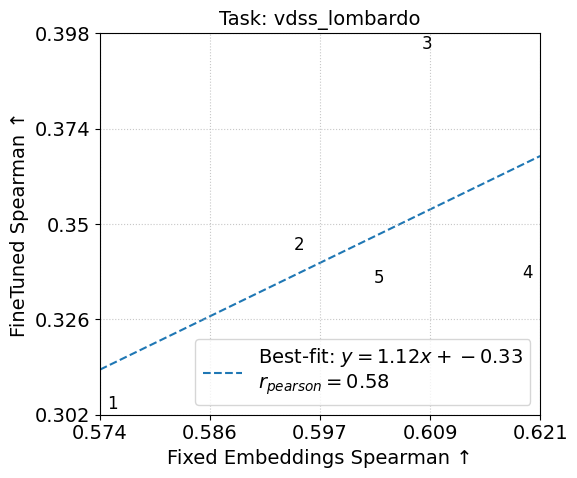}
  \end{subfigure}
  \caption{Frozen embeddings vs. finetuned performance for every layer of OrbV3-Direct.}
\end{figure}

\begin{figure}[!htpb]
  \centering
  \begin{subfigure}[b]{0.49\textwidth}
    \centering
    \includegraphics[width=\textwidth]{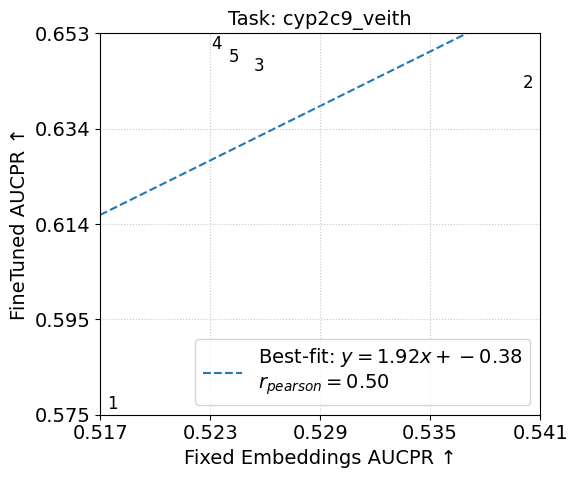}
  \end{subfigure}
  \hfill
  \begin{subfigure}[b]{0.49\textwidth}
    \centering
    \includegraphics[width=\textwidth]{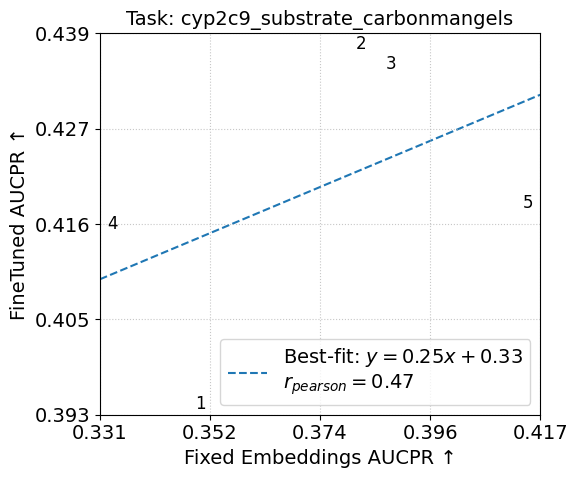}
  \end{subfigure}
  \caption{Frozen embeddings vs. finetuned performance for every layer of OrbV3-Direct.}
\end{figure}

\begin{figure}[!htpb]
  \centering
  \begin{subfigure}[b]{0.49\textwidth}
    \centering
    \includegraphics[width=\textwidth]{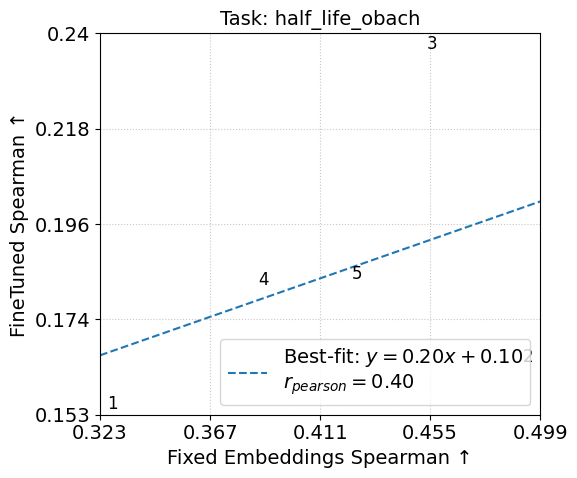}
  \end{subfigure}
  \hfill
  \begin{subfigure}[b]{0.49\textwidth}
    \centering
    \includegraphics[width=\textwidth]{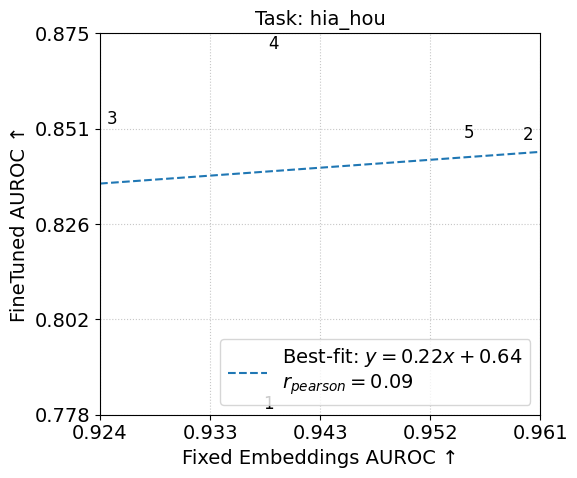}
  \end{subfigure}
  \caption{Frozen embeddings vs. finetuned performance for every layer of OrbV3-Direct.}
\end{figure}

\begin{figure}[!htpb]
  \centering
  \begin{subfigure}[b]{0.49\textwidth}
    \centering
    \includegraphics[width=\textwidth]{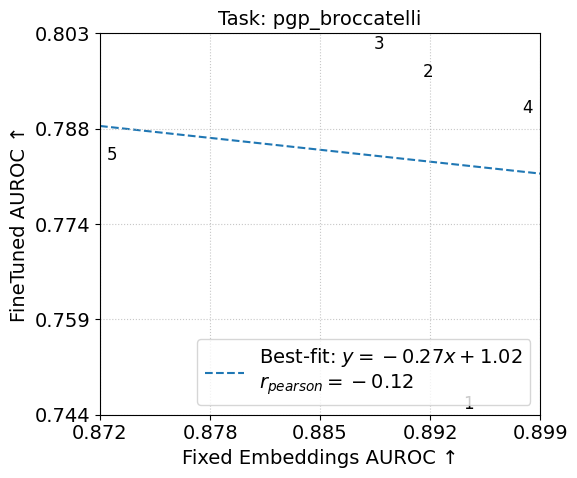}
  \end{subfigure}
  \hfill
  \begin{subfigure}[b]{0.49\textwidth}
    \centering
    \includegraphics[width=\textwidth]{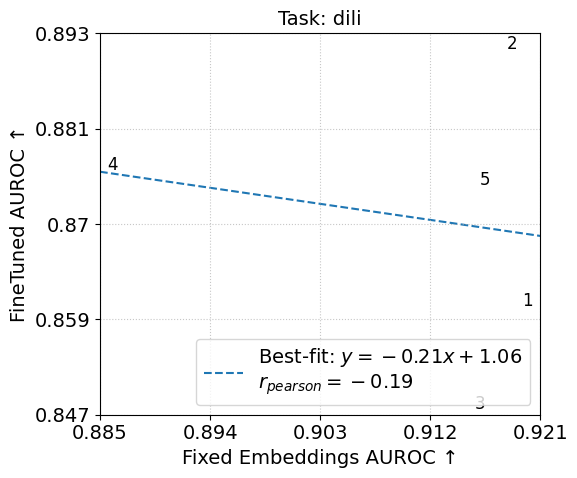}
  \end{subfigure}
  \caption{Frozen embeddings vs. finetuned performance for every layer of OrbV3-Direct.}
\end{figure}

\begin{figure}[!htpb]
  \centering
  \begin{subfigure}[b]{0.49\textwidth}
    \centering
    \includegraphics[width=\textwidth]{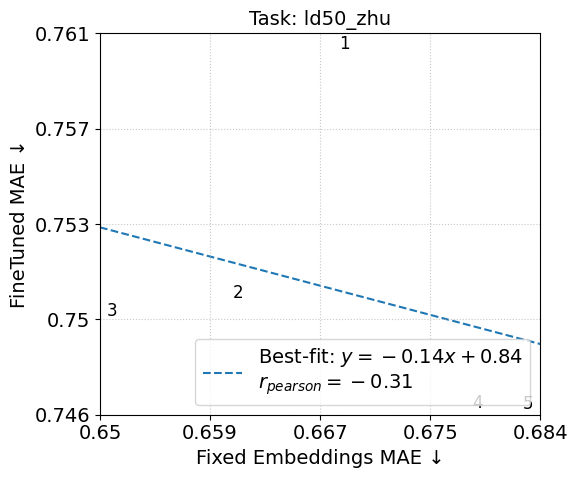}
  \end{subfigure}
  \hfill
  \begin{subfigure}[b]{0.49\textwidth}
    \centering
    \includegraphics[width=\textwidth]{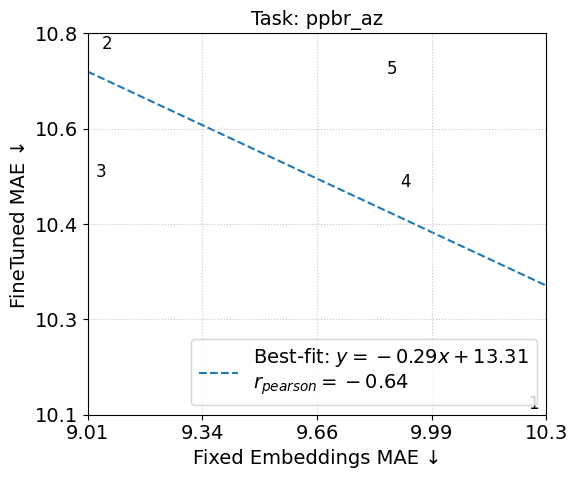}
  \end{subfigure}
  \caption{Frozen embeddings vs. finetuned performance for every layer of OrbV3-Direct.}
\end{figure}

\begin{figure}[!htpb]
  \centering
  \begin{subfigure}[b]{0.49\textwidth}
    \centering
    \includegraphics[width=\textwidth]{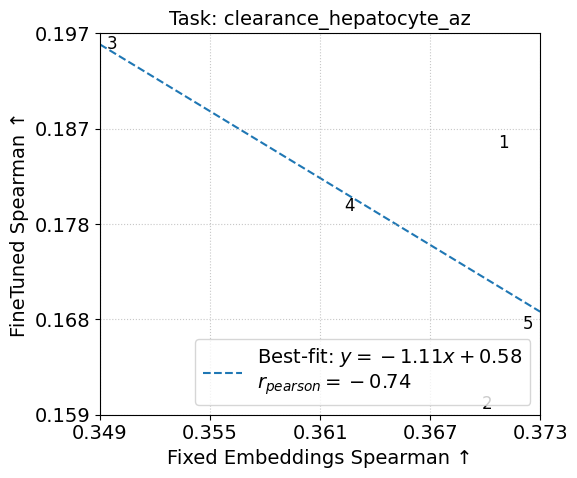}
  \end{subfigure}
  \hfill
  \begin{subfigure}[b]{0.49\textwidth}
    \centering
    \includegraphics[width=\textwidth]{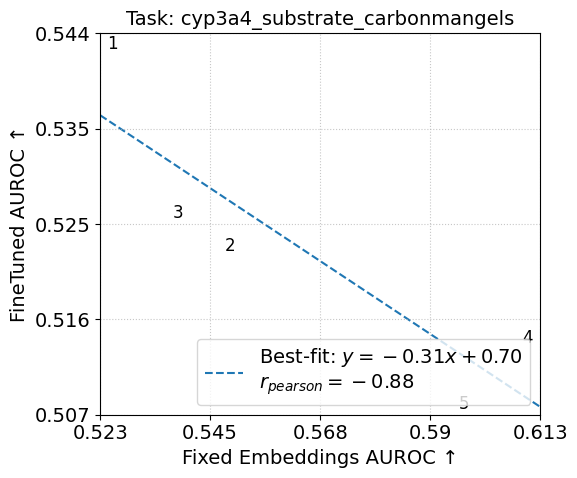}
  \end{subfigure}
  \caption{Frozen embeddings vs. finetuned performance for every layer of OrbV3-Direct.}
\end{figure}

\newpage

\section{OrbV3-Conservative: Frozen embedding vs. Finetuned Scatter Plots} \label{appendix G}

This appendix section presents a series of scatter plots dedicated to the OrbV3-Conservative model. Each plot illustrates the relationship between frozen embedding performance and full finetuning performance across the model's different layers for a specific downstream task. The plots were ordered by Pearson correlation.

\begin{figure}[!htpb]
  \centering
  \begin{subfigure}[b]{0.49\textwidth}
    \centering
    \includegraphics[width=\textwidth]{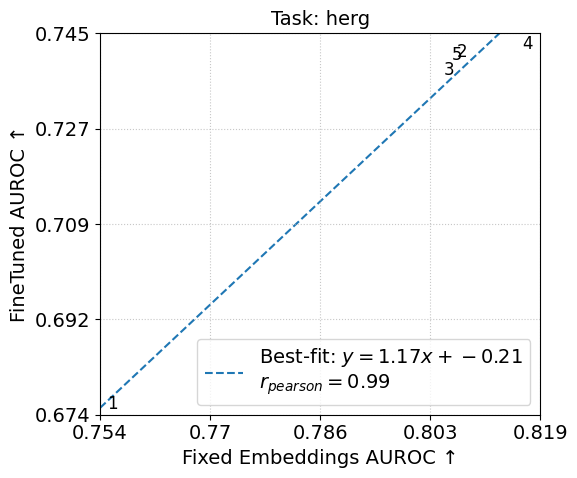}
  \end{subfigure}
  \hfill
  \begin{subfigure}[b]{0.49\textwidth}
    \centering
    \includegraphics[width=\textwidth]{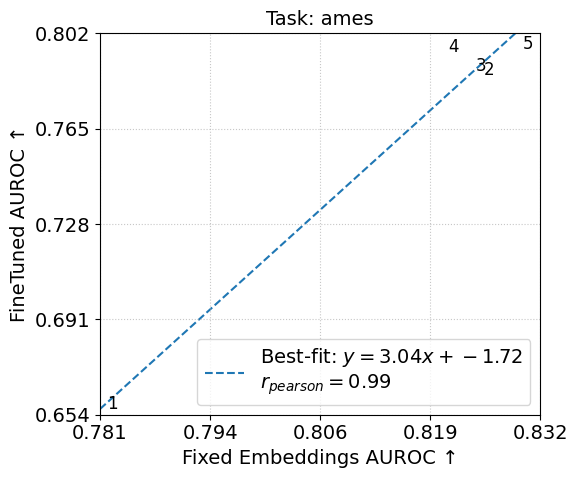}
  \end{subfigure}
  \caption{Frozen embeddings vs. finetuned performance for every layer of OrbV3-Conservative.}
\end{figure}

\begin{figure}[!htpb]
  \centering
  \begin{subfigure}[b]{0.49\textwidth}
    \centering
    \includegraphics[width=\textwidth]{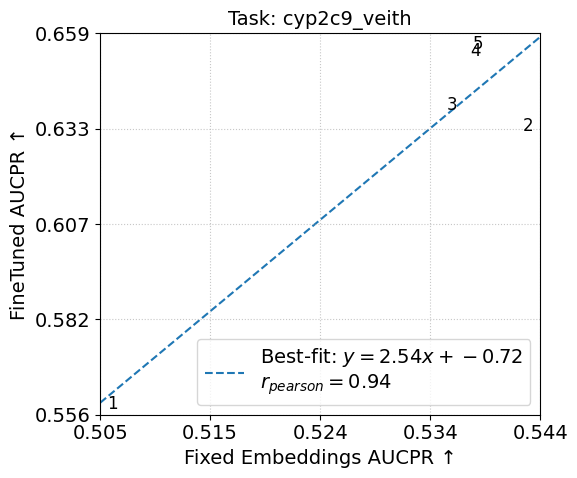}
  \end{subfigure}
  \hfill
  \begin{subfigure}[b]{0.49\textwidth}
    \centering
    \includegraphics[width=\textwidth]{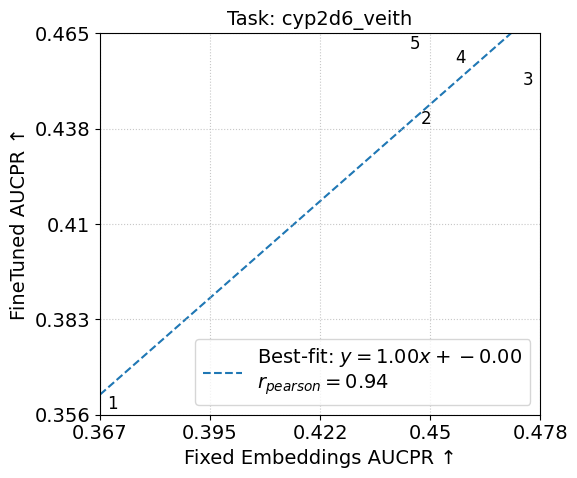}
  \end{subfigure}
  \caption{Frozen embeddings vs. finetuned performance for every layer of OrbV3-Conservative.}
\end{figure}

\begin{figure}[!htpb]
  \centering
  \begin{subfigure}[b]{0.49\textwidth}
    \centering
    \includegraphics[width=\textwidth]{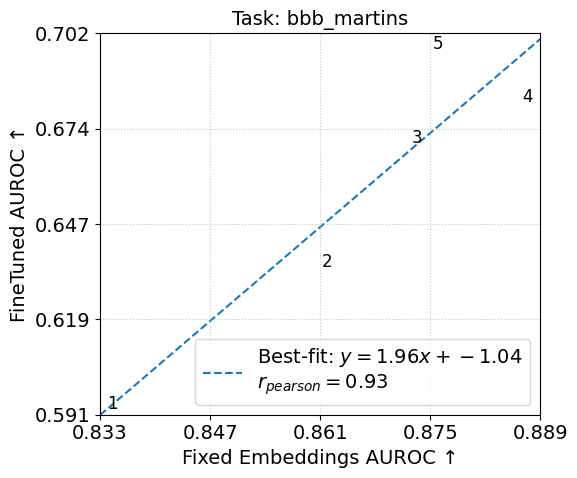}
  \end{subfigure}
  \hfill
  \begin{subfigure}[b]{0.49\textwidth}
    \centering
    \includegraphics[width=\textwidth]{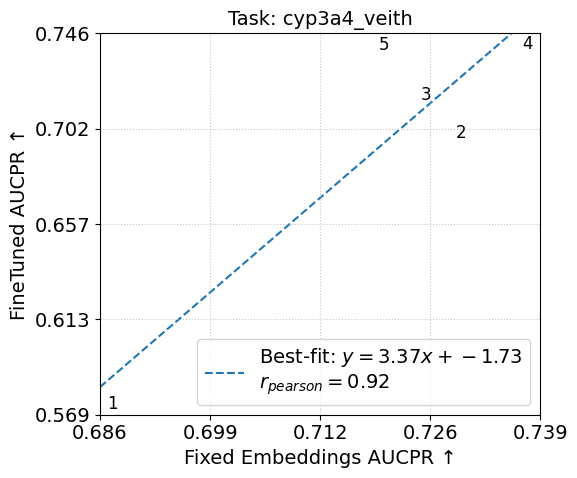}
  \end{subfigure}
  \caption{Frozen embeddings vs. finetuned performance for every layer of OrbV3-Conservative.}
\end{figure}

\begin{figure}[!htpb]
  \centering
  \begin{subfigure}[b]{0.49\textwidth}
    \centering
    \includegraphics[width=\textwidth]{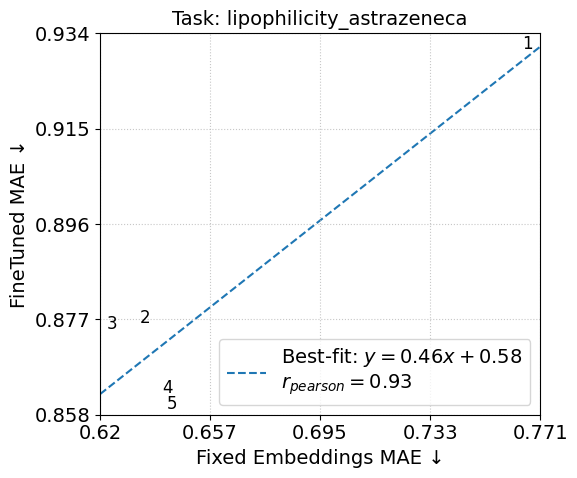}
  \end{subfigure}
  \hfill
  \begin{subfigure}[b]{0.49\textwidth}
    \centering
    \includegraphics[width=\textwidth]{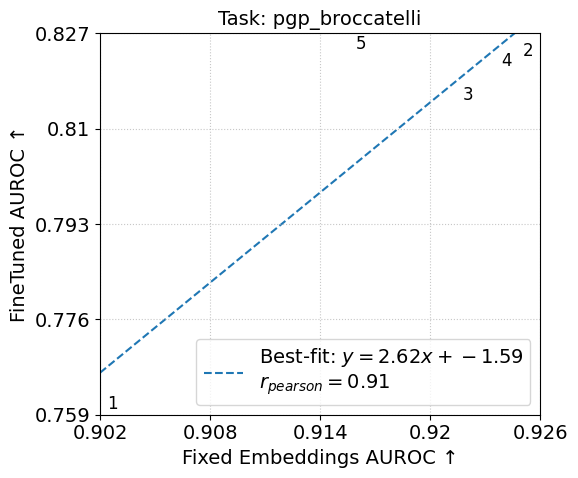}
  \end{subfigure}
  \caption{Frozen embeddings vs. finetuned performance for every layer of OrbV3-Conservative.}
\end{figure}

\begin{figure}[!htpb]
  \centering
  \begin{subfigure}[b]{0.49\textwidth}
    \centering
    \includegraphics[width=\textwidth]{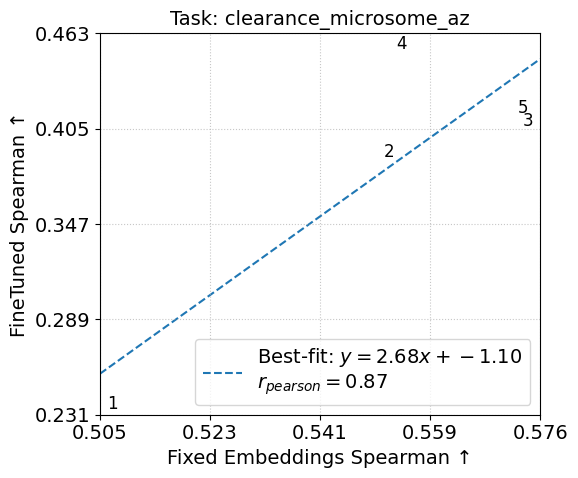}
  \end{subfigure}
  \hfill
  \begin{subfigure}[b]{0.49\textwidth}
    \centering
    \includegraphics[width=\textwidth]{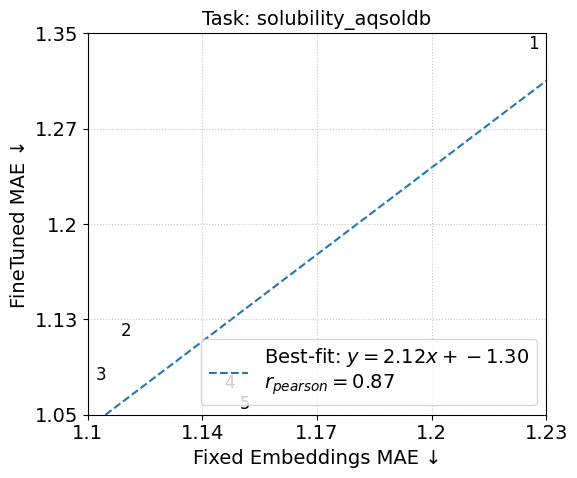}
  \end{subfigure}
  \caption{Frozen embeddings vs. finetuned performance for every layer of OrbV3-Conservative.}
\end{figure}

\begin{figure}[!htpb]
  \centering
  \begin{subfigure}[b]{0.49\textwidth}
    \centering
    \includegraphics[width=\textwidth]{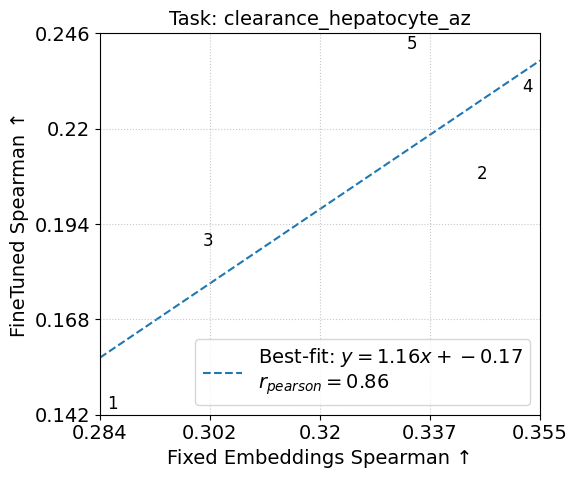}
  \end{subfigure}
  \hfill
  \begin{subfigure}[b]{0.49\textwidth}
    \centering
    \includegraphics[width=\textwidth]{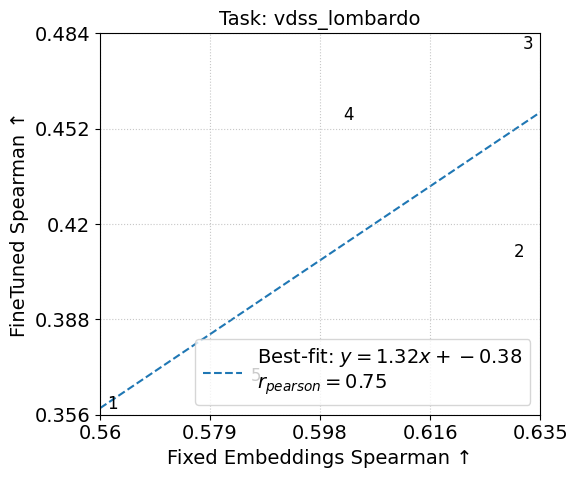}
  \end{subfigure}
  \caption{Frozen embeddings vs. finetuned performance for every layer of OrbV3-Conservative.}
\end{figure}

\begin{figure}[!htpb]
  \centering
  \begin{subfigure}[b]{0.49\textwidth}
    \centering
    \includegraphics[width=\textwidth]{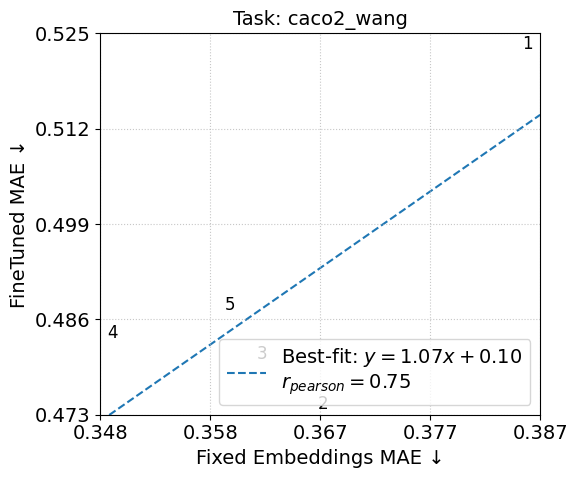}
  \end{subfigure}
  \hfill
  \begin{subfigure}[b]{0.49\textwidth}
    \centering
    \includegraphics[width=\textwidth]{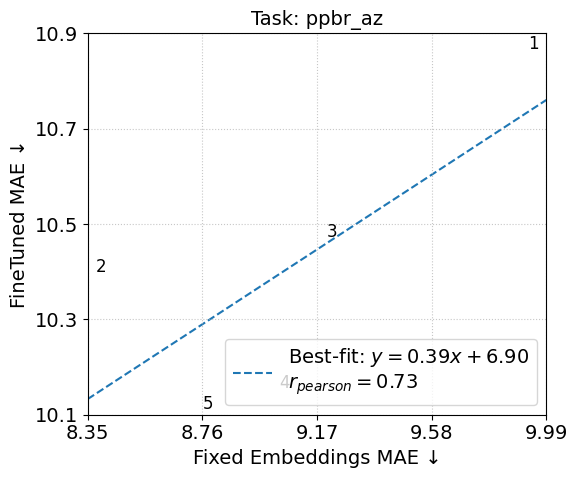}
  \end{subfigure}
  \caption{Frozen embeddings vs. finetuned performance for every layer of OrbV3-Conservative.}
\end{figure}

\begin{figure}[!htpb]
  \centering
  \begin{subfigure}[b]{0.49\textwidth}
    \centering
    \includegraphics[width=\textwidth]{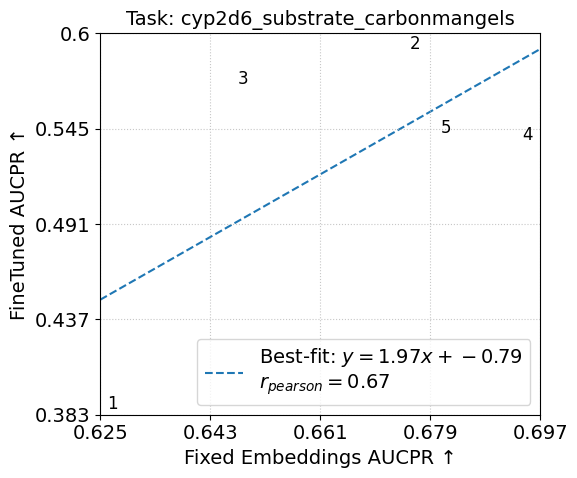}
  \end{subfigure}
  \hfill
  \begin{subfigure}[b]{0.49\textwidth}
    \centering
    \includegraphics[width=\textwidth]{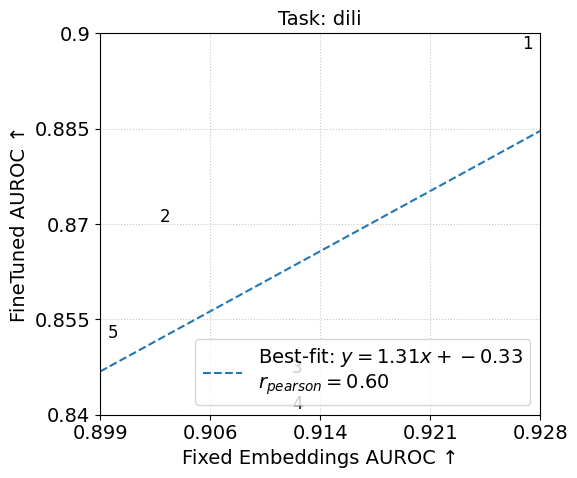}
  \end{subfigure}
  \caption{Frozen embeddings vs. finetuned performance for every layer of OrbV3-Conservative.}
\end{figure}

\begin{figure}[!htpb]
  \centering
  \begin{subfigure}[b]{0.49\textwidth}
    \centering
    \includegraphics[width=\textwidth]{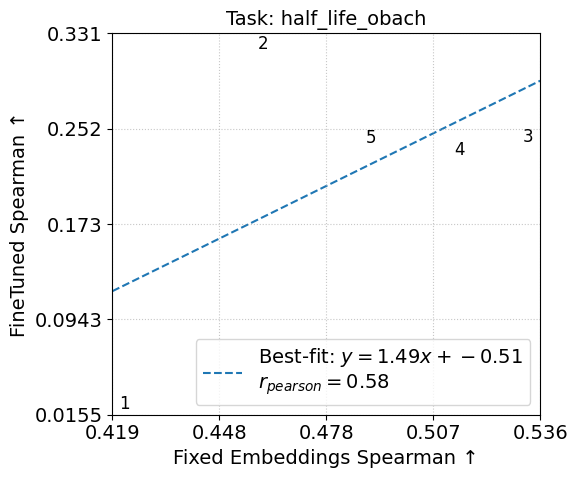}
  \end{subfigure}
  \hfill
  \begin{subfigure}[b]{0.49\textwidth}
    \centering
    \includegraphics[width=\textwidth]{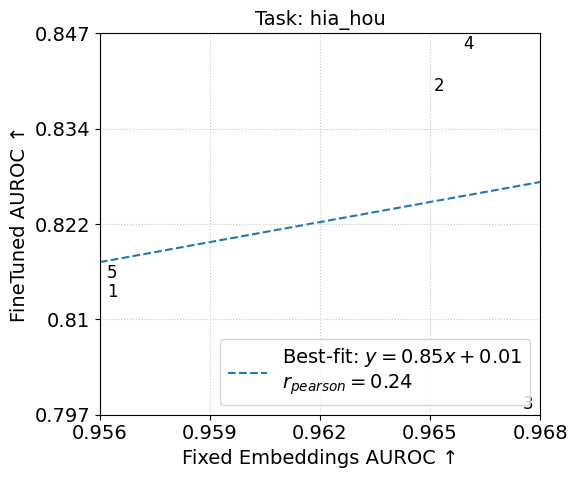}
  \end{subfigure}
  \caption{Frozen embeddings vs. finetuned performance for every layer of OrbV3-Conservative.}
\end{figure}

\begin{figure}[!htpb]
  \centering
  \begin{subfigure}[b]{0.49\textwidth}
    \centering
    \includegraphics[width=\textwidth]{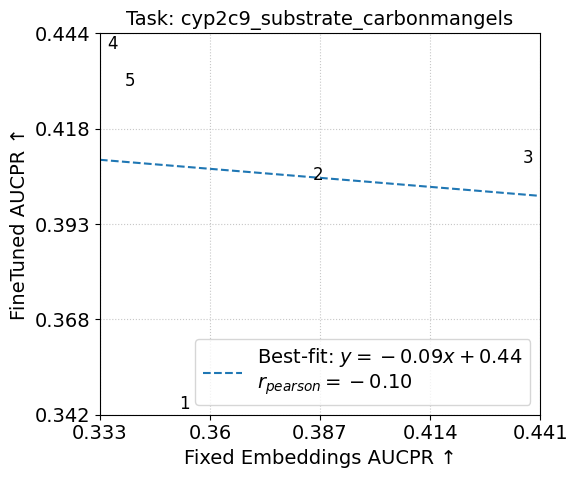}
  \end{subfigure}
  \hfill
  \begin{subfigure}[b]{0.49\textwidth}
    \centering
    \includegraphics[width=\textwidth]{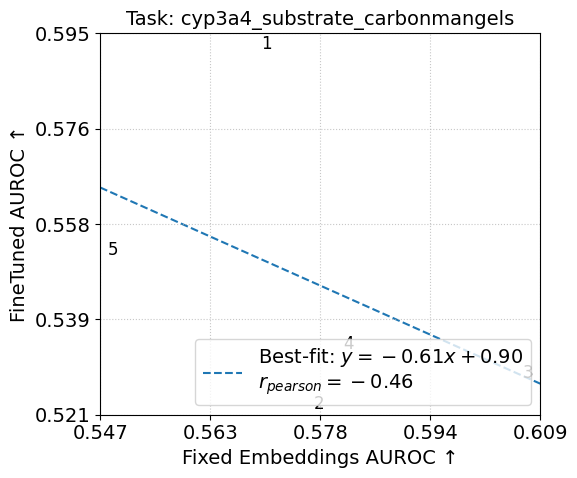}
  \end{subfigure}
  \caption{Frozen embeddings vs. finetuned performance for every layer of OrbV3-Conservative.}
\end{figure}

\begin{figure}[!htpb]
  \centering
  \begin{subfigure}[b]{0.49\textwidth}
    \centering
    \includegraphics[width=\textwidth]{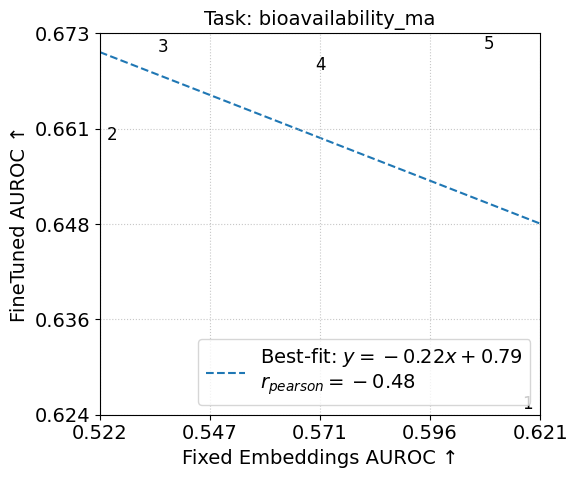}
  \end{subfigure}
  \hfill
  \begin{subfigure}[b]{0.49\textwidth}
    \centering
    \includegraphics[width=\textwidth]{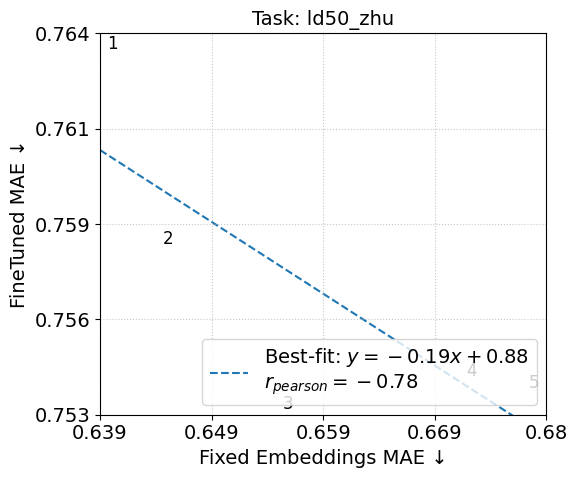}
  \end{subfigure}
  \caption{Frozen embeddings vs. finetuned performance for every layer of OrbV3-Conservative.}
\end{figure}

\newpage
\section{Pos-EGNN: Frozen embedding vs. Finetuned Scatter Plots} \label{appendix H}

This appendix section presents a series of scatter plots dedicated to the Pos-EGNN model. Each plot illustrates the relationship between frozen embedding performance and full finetuning performance across the model's different layers for a specific downstream task. The plots were ordered by Pearson correlation.

\begin{figure}[!htpb]
  \centering
  \begin{subfigure}[b]{0.49\textwidth}
    \centering
    \includegraphics[width=\textwidth]{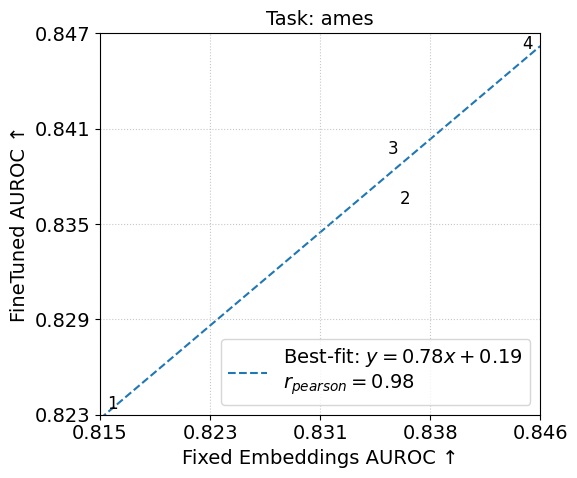}
  \end{subfigure}
  \hfill
  \begin{subfigure}[b]{0.49\textwidth}
    \centering
    \includegraphics[width=\textwidth]{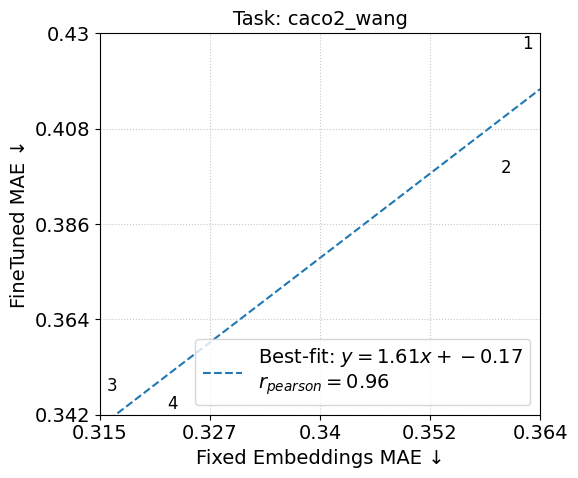}
  \end{subfigure}
  \caption{Frozen embeddings vs. finetuned performance for every layer of Pos-EGNN.}
\end{figure}

\begin{figure}[!htpb]
  \centering
  \begin{subfigure}[b]{0.49\textwidth}
    \centering
    \includegraphics[width=\textwidth]{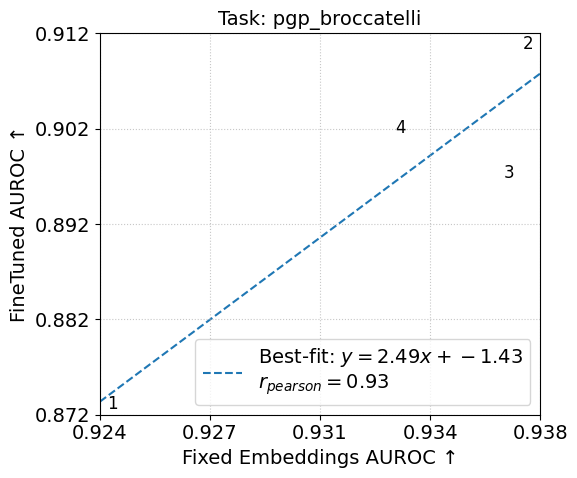}
  \end{subfigure}
  \hfill
  \begin{subfigure}[b]{0.49\textwidth}
    \centering
    \includegraphics[width=\textwidth]{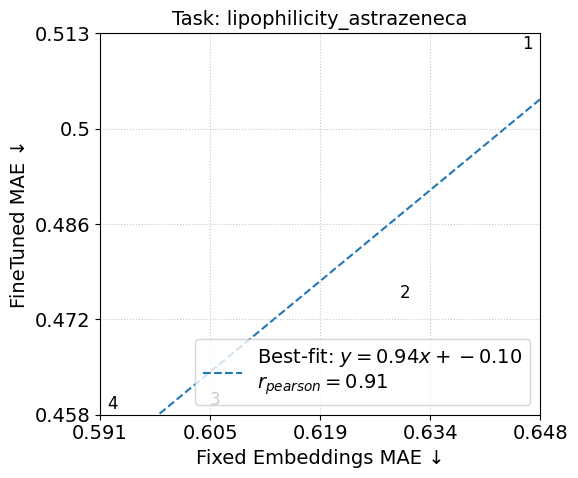}
  \end{subfigure}
  \caption{Frozen embeddings vs. finetuned performance for every layer of Pos-EGNN.}
\end{figure}

\begin{figure}[!htpb]
  \centering
  \begin{subfigure}[b]{0.49\textwidth}
    \centering
    \includegraphics[width=\textwidth]{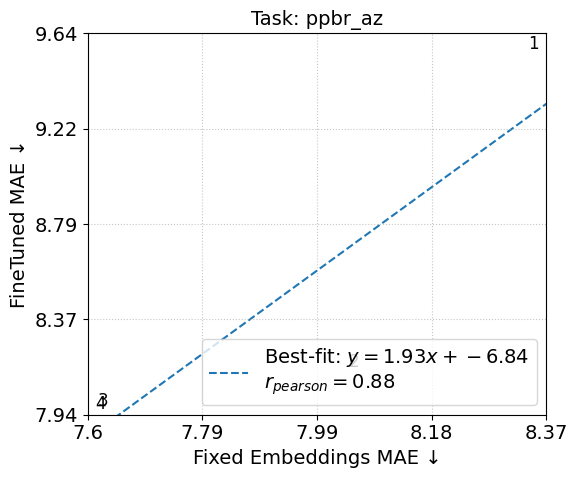}
  \end{subfigure}
  \hfill
  \begin{subfigure}[b]{0.49\textwidth}
    \centering
    \includegraphics[width=\textwidth]{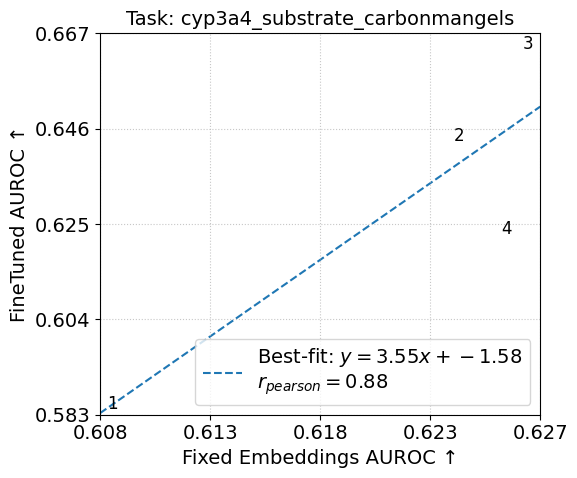}
  \end{subfigure}
  \caption{Frozen embeddings vs. finetuned performance for every layer of Pos-EGNN.}
\end{figure}

\begin{figure}[!htpb]
  \centering
  \begin{subfigure}[b]{0.49\textwidth}
    \centering
    \includegraphics[width=\textwidth]{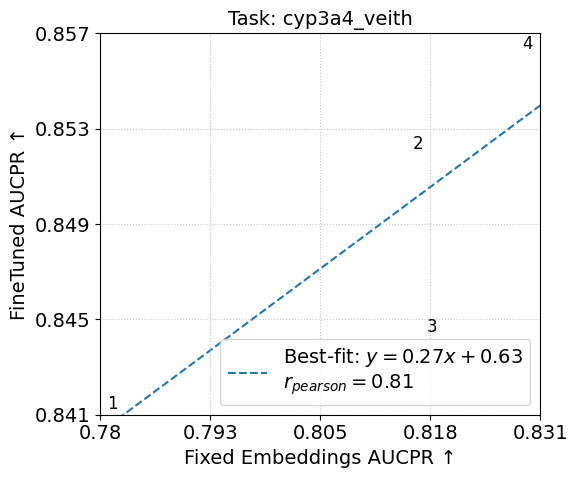}
  \end{subfigure}
  \hfill
  \begin{subfigure}[b]{0.49\textwidth}
    \centering
    \includegraphics[width=\textwidth]{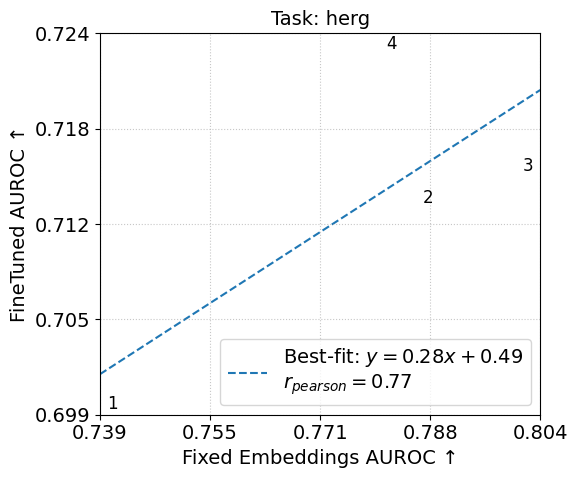}
  \end{subfigure}
  \caption{Frozen embeddings vs. finetuned performance for every layer of Pos-EGNN.}
\end{figure}

\begin{figure}[!htpb]
  \centering
  \begin{subfigure}[b]{0.49\textwidth}
    \centering
    \includegraphics[width=\textwidth]{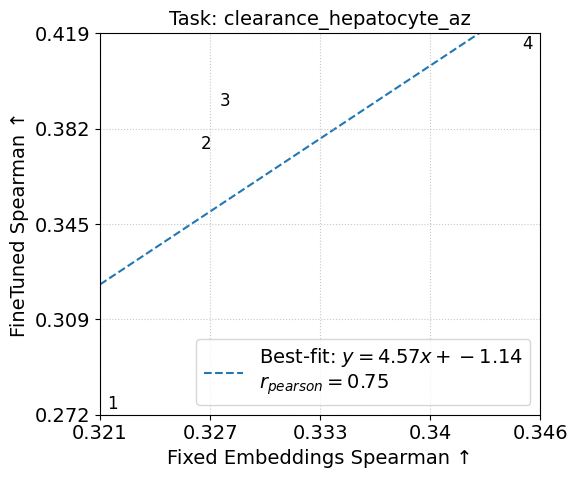}
  \end{subfigure}
  \hfill
  \begin{subfigure}[b]{0.49\textwidth}
    \centering
    \includegraphics[width=\textwidth]{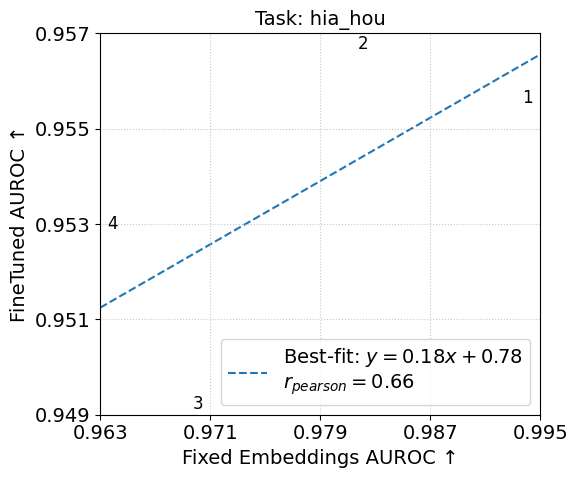}
  \end{subfigure}
  \caption{Frozen embeddings vs. finetuned performance for every layer of Pos-EGNN.}
\end{figure}

\begin{figure}[!htpb]
  \centering
  \begin{subfigure}[b]{0.49\textwidth}
    \centering
    \includegraphics[width=\textwidth]{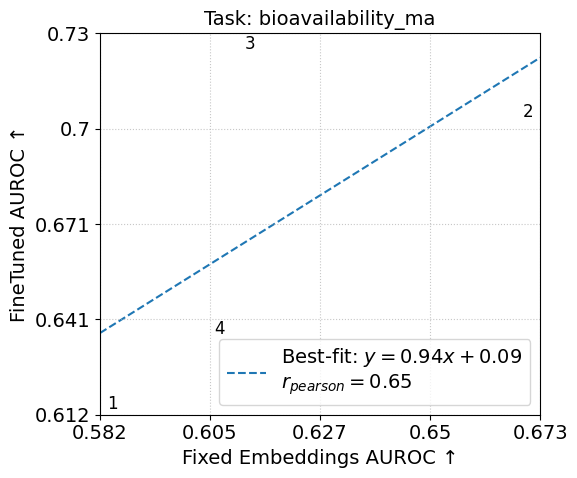}
  \end{subfigure}
  \hfill
  \begin{subfigure}[b]{0.49\textwidth}
    \centering
    \includegraphics[width=\textwidth]{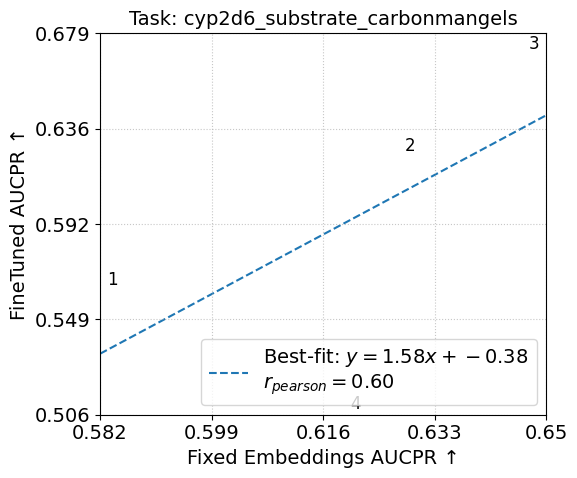}
  \end{subfigure}
  \caption{Frozen embeddings vs. finetuned performance for every layer of Pos-EGNN.}
\end{figure}

\begin{figure}[!htpb]
  \centering
  \begin{subfigure}[b]{0.49\textwidth}
    \centering
    \includegraphics[width=\textwidth]{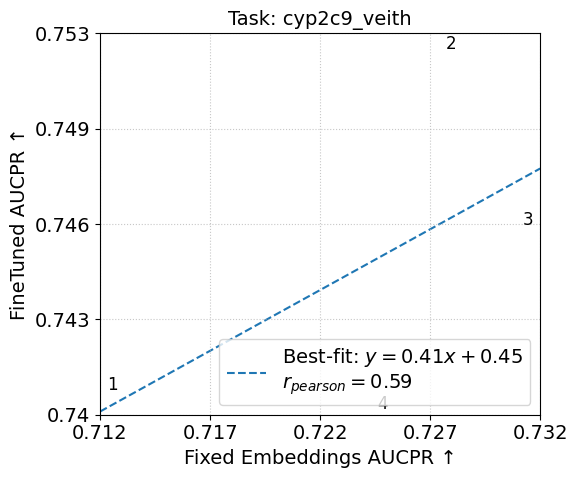}
  \end{subfigure}
  \hfill
  \begin{subfigure}[b]{0.49\textwidth}
    \centering
    \includegraphics[width=\textwidth]{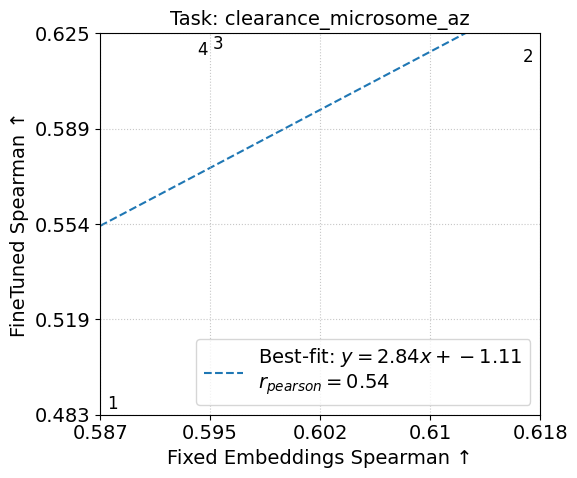}
  \end{subfigure}
  \caption{Frozen embeddings vs. finetuned performance for every layer of Pos-EGNN.}
\end{figure}

\begin{figure}[!htpb]
  \centering
  \begin{subfigure}[b]{0.49\textwidth}
    \centering
    \includegraphics[width=\textwidth]{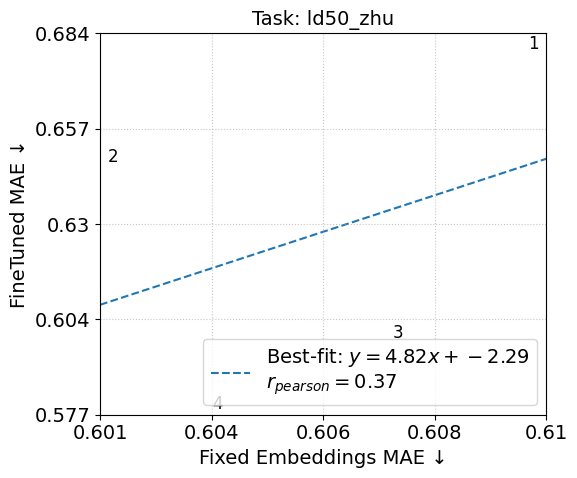}
  \end{subfigure}
  \hfill
  \begin{subfigure}[b]{0.49\textwidth}
    \centering
    \includegraphics[width=\textwidth]{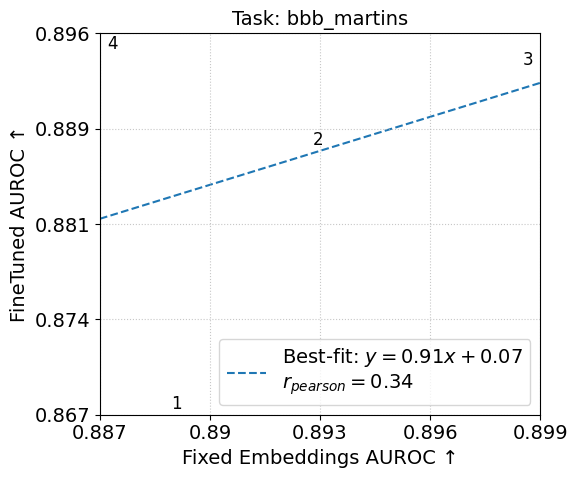}
  \end{subfigure}
  \caption{Frozen embeddings vs. finetuned performance for every layer of Pos-EGNN.}
\end{figure}

\begin{figure}[!htpb]
  \centering
  \begin{subfigure}[b]{0.49\textwidth}
    \centering
    \includegraphics[width=\textwidth]{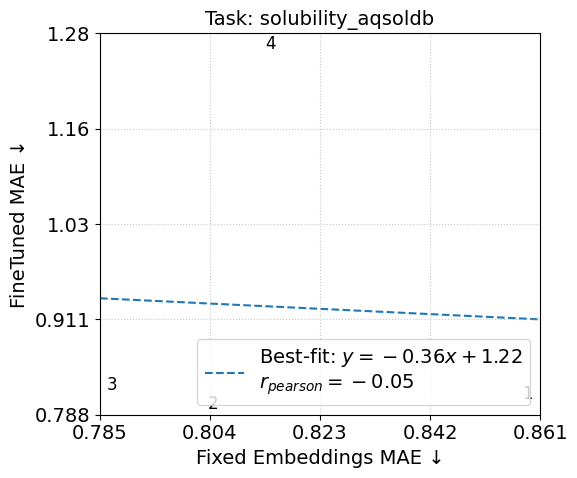}
  \end{subfigure}
  \hfill
  \begin{subfigure}[b]{0.49\textwidth}
    \centering
    \includegraphics[width=\textwidth]{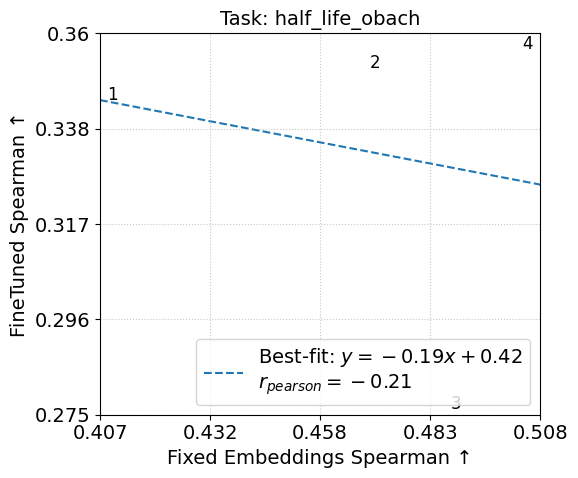}
  \end{subfigure}
  \caption{Frozen embeddings vs. finetuned performance for every layer of Pos-EGNN.}
\end{figure}

\begin{figure}[!htpb]
  \centering
  \begin{subfigure}[b]{0.49\textwidth}
    \centering
    \includegraphics[width=\textwidth]{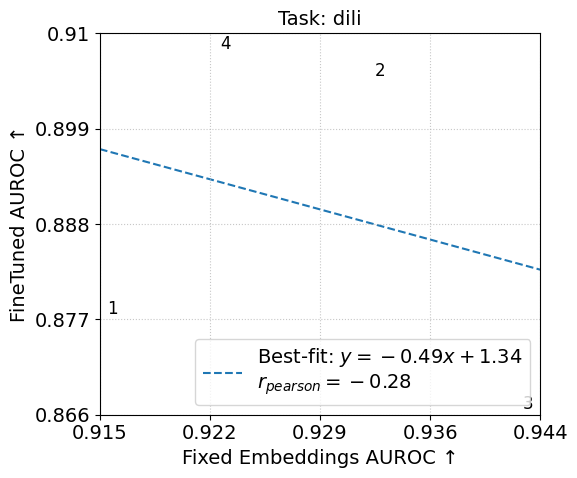}
  \end{subfigure}
  \hfill
  \begin{subfigure}[b]{0.49\textwidth}
    \centering
    \includegraphics[width=\textwidth]{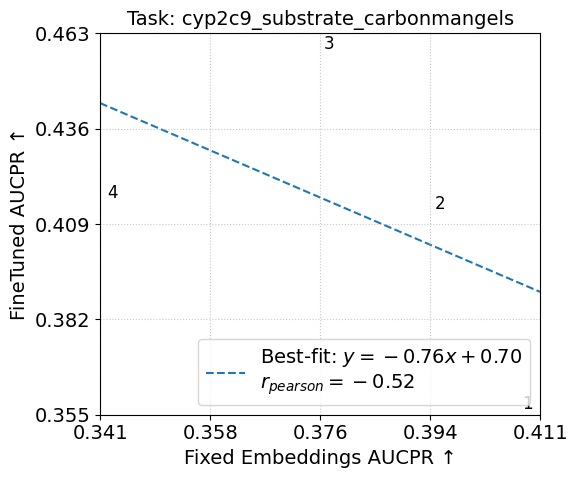}
  \end{subfigure}
  \caption{Frozen embeddings vs. finetuned performance for every layer of Pos-EGNN.}
\end{figure}

\begin{figure}[!htpb]
  \centering
  \begin{subfigure}[b]{0.49\textwidth}
    \centering
    \includegraphics[width=\textwidth]{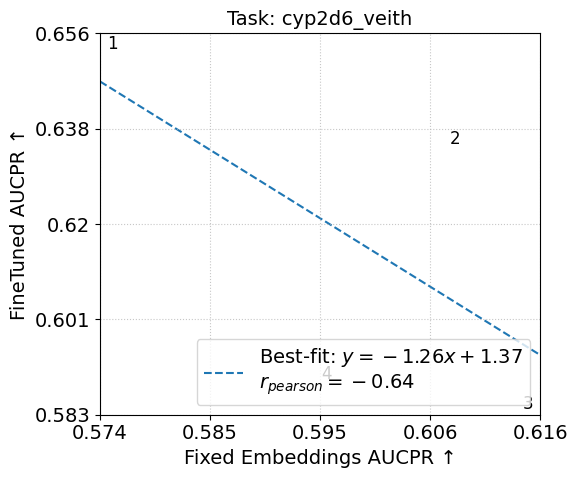}
  \end{subfigure}
  \hfill
  \begin{subfigure}[b]{0.49\textwidth}
    \centering
    \includegraphics[width=\textwidth]{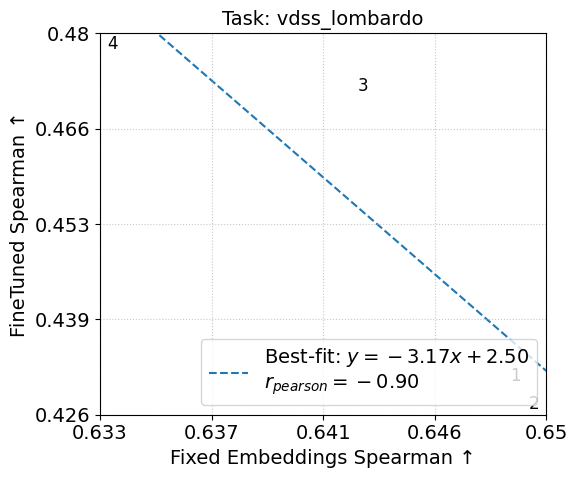}
  \end{subfigure}
  \caption{Frozen embeddings vs. finetuned performance for every layer of Pos-EGNN.}
\end{figure}

\end{document}